\documentclass[10pt,twocolumn,letterpaper]{article}

\usepackage{cvpr}
\usepackage{times}
\usepackage{epsfig}
\usepackage{graphicx}
\usepackage{amsmath}
\usepackage{amssymb}
\usepackage{multirow}
\usepackage{makecell}
\usepackage{colortbl}

\definecolor{gray}{gray}{0.7}

\usepackage{multirow}

\usepackage{booktabs}

\usepackage{pifont}

\makeatletter
\renewcommand\paragraph{\@startsection{paragraph}{4}{\z@}%
  {.5em \@plus1ex \@minus.2ex}%
  {-1em}%
  {\normalfont\normalsize\bfseries}}
\makeatother

\definecolor{larry_color}{rgb}{0.2,.64,0}
\definecolor{ross_color}{rgb}{.6,.25,.5}
\definecolor{bh_color}{rgb}{.1,.1,.8}
\definecolor{todo_color}{rgb}{.8,.3,.1}
\definecolor{new_color}{rgb}{.1,.7,.8}

\definecolor{red}{rgb}{0.68,0.14,0.14}
\definecolor{blue}{rgb}{0.16,0.29,0.84}
\definecolor{green}{rgb}{0.11,0.41,0.08}
\definecolor{purple}{rgb}{0.5,0.15,0.75}
\definecolor{orange}{rgb}{1,0.57,0.2}
\definecolor{brown}{rgb}{0.5,0.29,0.1}

\newenvironment{packed_item}{
\begin{itemize}
  \setlength{\itemsep}{1pt}
  \setlength{\parskip}{0pt}
  \setlength{\parsep}{0pt}
}{\end{itemize}}

\newcommand{\leaveout}[1]{}

\usepackage{amsopn}

\usepackage[pagebackref=true,breaklinks=true,letterpaper=true,colorlinks,bookmarks=false]{hyperref}
\usepackage{colortbl}
\definecolor{gray}{gray}{0.7}

\cvprfinalcopy 

\def\cvprPaperID{1062} 

\begin{document}

\title{CLEVR: A Diagnostic Dataset for\\Compositional Language and Elementary Visual Reasoning}

\author{
  \begin{minipage}{0.3\textwidth}
    \centering
    Justin Johnson\textsuperscript{1,2}\footnotemark \\*
    Li Fei-Fei\textsuperscript{1}
  \end{minipage}
  \begin{minipage}{0.3\textwidth}
    \centering
    Bharath Hariharan\textsuperscript{2} \\*
    C. Lawrence Zitnick\textsuperscript{2}
  \end{minipage}
  \begin{minipage}{0.3\textwidth}
    \centering
    Laurens van der Maaten\textsuperscript{2} \\*
    Ross Girshick\textsuperscript{2}
  \end{minipage}
  \\*[14pt]
  \begin{minipage}{0.4\textwidth}
    \centering
    \textsuperscript{1}Stanford University
  \end{minipage}
  \begin{minipage}{0.4\textwidth}
    \centering
    \textsuperscript{2}Facebook AI Research
  \end{minipage}
}

\maketitle
\renewcommand*{\thefootnote}{\fnsymbol{footnote}}
\setcounter{footnote}{1}
\footnotetext{Work done during an internship at FAIR.}
\renewcommand*{\thefootnote}{\arabic{footnote}}
\setcounter{footnote}{0}

\begin{abstract}
When building artificial intelligence systems that can reason and answer questions about visual data, we need diagnostic tests to analyze our progress and discover shortcomings. Existing benchmarks for visual question answering can help, but have strong biases that models can exploit to correctly answer questions without reasoning. They also conflate multiple sources of error, making it hard to pinpoint model weaknesses. We present a diagnostic dataset that tests a range of visual reasoning abilities. It contains minimal biases and has detailed annotations describing the kind of reasoning each question requires. We use this dataset to analyze a variety of modern visual reasoning systems, providing novel insights into their abilities and limitations.

\end{abstract}

\section{Introduction}
\label{Introduction}

A long-standing goal of artificial intelligence research is to develop systems that can reason and answer questions about visual information. Recently, several datasets have been introduced to study this problem \cite{antol15,gao15,krishna16,malinowski14,ren15,yu15,zhu15}. Each of these Visual Question Answering (VQA) datasets contains challenging natural language questions about images.
Correctly answering these questions requires perceptual abilities such as recognizing objects, attributes, and spatial relationships as well as higher-level skills such as counting, performing logical inference, making comparisons, or leveraging commonsense world knowledge~\cite{ray16}.
Numerous methods have attacked these problems \cite{andreas16b,andreas16,fukui16,hiatt16,yang16}, but many show only marginal improvements over strong baselines \cite{antol15,jabri16,zhou15}. Unfortunately, our ability to understand the limitations of these methods is impeded by the inherent complexity of the VQA task. Are methods hampered by failures in recognition, poor reasoning, lack of commonsense knowledge, or something else?

\begin{figure}
  \centering
  \includegraphics[width=0.42\textwidth]{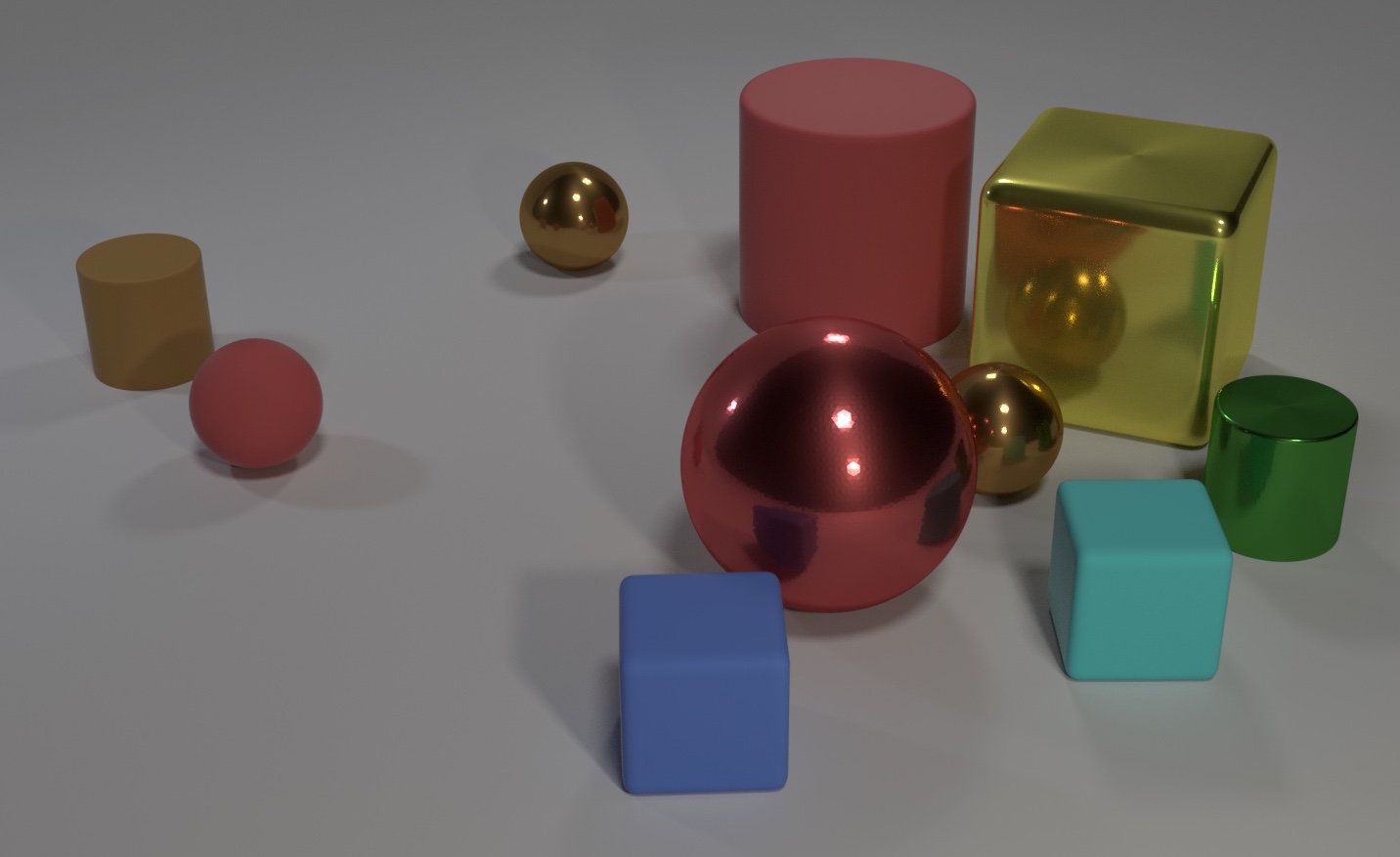} \\*
  \vspace{1mm}
  \begin{minipage}{0.445\textwidth}
    \footnotesize
    \textbf{Q:} Are there an \textcolor{blue}{equal} \textcolor{red}{number} of
    \textcolor{brown}{large} things and \textcolor{brown}{metal spheres}? \\*
    \textbf{Q:} \textcolor{brown}{What size} is the \textcolor{brown}{cylinder}
    \textcolor{green}{that is left of} the \textcolor{brown}{brown metal} thing
    \textcolor{green}{that is left of} the \textcolor{brown}{big sphere}?
    \textbf{Q:} There is a \textcolor{brown}{sphere} with the \textcolor{blue}{same size as}
    the \textcolor{brown}{metal cube}; is it
    \textcolor{blue}{made of the same material as} the \textcolor{brown}{small red sphere}? \\*
    \textbf{Q:} \textcolor{red}{How many} objects \textcolor{purple}{are either}
    \textcolor{brown}{small cylinders}
    \textcolor{purple}{or} \textcolor{brown}{metal} things?
  \end{minipage}
  \vspace{1mm}
  \caption{A sample image and questions from CLEVR. Questions test aspects of visual reasoning
    such as \textcolor{brown}{attribute identification}, \textcolor{red}{counting},
    \textcolor{blue}{comparison}, \textcolor{green}{multiple attention},
    and \textcolor{purple}{logical operations}.
  }
  \vspace{-4mm}
  \label{fig:teaser}
\end{figure}

The difficulty of understanding a system's competences is exemplified by \emph{Clever Hans}, a 1900s era horse who appeared to be able to answer arithmetic questions. Careful observation revealed that Hans was correctly ``answering'' questions by reacting to cues read off his human observers \cite{pfungst11}. Statistical learning systems, like those used for VQA, may develop similar ``cheating'' approaches to superficially ``solve'' tasks without learning the underlying reasoning processes~\cite{sturm14,sturm16}.
For instance, a statistical learner may correctly answer the question \emph{``What covers the ground?''} not because it understands the scene
but because biased datasets often ask questions about the \emph{ground} when it is snow-covered~\cite{agrawal16,zhang16}. How can we determine whether a system is capable of sophisticated reasoning and not just exploiting biases of the world, similar to Clever Hans?

In this paper we propose a \emph{diagnostic dataset} for studying the ability of VQA systems to perform visual reasoning. We refer to this dataset as the Compositional Language and Elementary Visual Reasoning diagnostics dataset (CLEVR; pronounced as \emph{clever} in homage to Hans). CLEVR contains 100k rendered images and about one million automatically-generated questions, of which 853k are unique. It has challenging images and questions that test visual reasoning abilities such as counting, comparing, logical reasoning, and storing information in memory, as illustrated in Figure \ref{fig:teaser}.

We designed CLEVR with the explicit goal of enabling detailed analysis of visual reasoning.
Our images depict simple 3D shapes; this simplifies recognition and allows us to focus on \emph{reasoning skills}.
We ensure that the information in each image is \emph{complete and exclusive} so that external information sources, such as commonsense knowledge, cannot increase the chance of correctly answering questions.
We minimize question-conditional bias via rejection sampling within families of related questions, and avoid degenerate questions that are seemingly complex but contain simple shortcuts to the correct answer. 
Finally, we use structured ground-truth representations for both images and questions: images are annotated with ground-truth object positions and attributes, and questions are represented as \emph{functional programs} that can be executed to answer the question (see Section~\ref{sec:question-gen}). These representations facilitate in-depth analyses not possible with traditional VQA datasets.

These design choices also mean that while images in CLEVR may be visually simple, its questions are complex and require a range of reasoning skills.
For instance, factorized representations may be required to generalize to unseen combinations of objects and attributes. 
Tasks such as counting or comparing may require short-term memory \cite{hochreiter97} or attending to specific objects \cite{hiatt16,yang16}. 
Questions that combine multiple subtasks in diverse ways may require compositional systems~\cite{andreas16b,andreas16} to answer.

We use CLEVR to analyze a suite of VQA models and discover weaknesses that are not widely known.
For example, we find that current state-of-the-art VQA models struggle on tasks requiring \emph{short term memory}, such as comparing the attributes of objects, or \emph{compositional reasoning}, such as recognizing novel attribute combinations.
These observations point to novel avenues for further research.

Finally, we stress that accuracy on CLEVR is not an end goal in itself: a hand-crafted system with explicit knowledge of the CLEVR universe might work well, but will not generalize to real-world settings.
Therefore CLEVR should be used \emph{in conjunction} with other VQA datasets in order to study the reasoning abilities of general VQA systems.

The CLEVR dataset, as well as code for generating new images and questions,
will be made publicly available.

\section{Related Work}
\label{Related Work}
In recent years, a range of benchmarks for visual understanding have been proposed, including datasets for image captioning \cite{chen15,farhadi10,lin14,young14}, referring to objects \cite{kazemzadeh14}, relational graph prediction \cite{krishna16}, and visual Turing tests \cite{geman15,malinowski14b}. CLEVR, our diagnostic dataset, is most closely related to benchmarks for visual question answering \cite{antol15,gao15,krishna16,malinowski14,ren15,TapaswiCVPR16,yu15,zhu15}, as it involves answering natural-language questions about images. The two main differences between CLEVR and other VQA datasets are that: (1) CLEVR minimizes biases of prior VQA datasets that can be used by learning systems to answer questions correctly without visual reasoning and (2) CLEVR's synthetic nature and detailed annotations facilitate in-depth analyses of reasoning abilities that are impossible with existing datasets.

Prior work has attempted to mitigate biases in VQA datasets in simple cases such as yes/no questions \cite{geman15,zhang16}, but it is difficult to apply such bias-reduction approaches to more complex questions without a high-quality semantic representation of both questions and answers. In CLEVR, this semantic representation is provided by the functional program underlying each image-question pair, and biases are largely eliminated via sampling. Winograd schemas~\cite{levesque11} are another approach for controlling bias in question answering: these questions are carefully designed to be ambiguous based on syntax alone and require commonsense knowledge. Unfortunately this approach does not scale gracefully: the first phase of the 2016 Winograd Schema Challenge consists of just 60 hand-designed questions.
CLEVR is also related to the bAbI question answering tasks~\cite{weston16} in that it aims to diagnose a set of clearly defined competences of a system, but CLEVR focuses on visual reasoning whereas bAbI is purely textual.

We are also not the first to consider synthetic data for studying (visual) reasoning. SHRDLU performed simple, interactive visual reasoning with the goal of moving specific objects in the visual scene \cite{winograd72}; this study was one of the first to demonstrate the brittleness of manually programmed semantic understanding. The pioneering DAQUAR dataset~\cite{malinowski15} contains both synthetic and human-written questions, but they only generate 420 synthetic questions using eight text templates. VQA~\cite{antol15} contains 150,000 natural-language questions about abstract scenes \cite{zitnick13}, but these questions do not control for question-conditional bias and are not equipped with functional program representations. CLEVR is similar in spirit to the SHAPES dataset \cite{andreas16}, but it is more complex and varied both in terms of visual content and question variety and complexity: SHAPES contains 15,616 total questions with just 244 unique questions while CLEVR contains nearly a million questions of which 853,554 are unique.

\begin{figure*}
  \centering
  \includegraphics[width=0.98\textwidth]{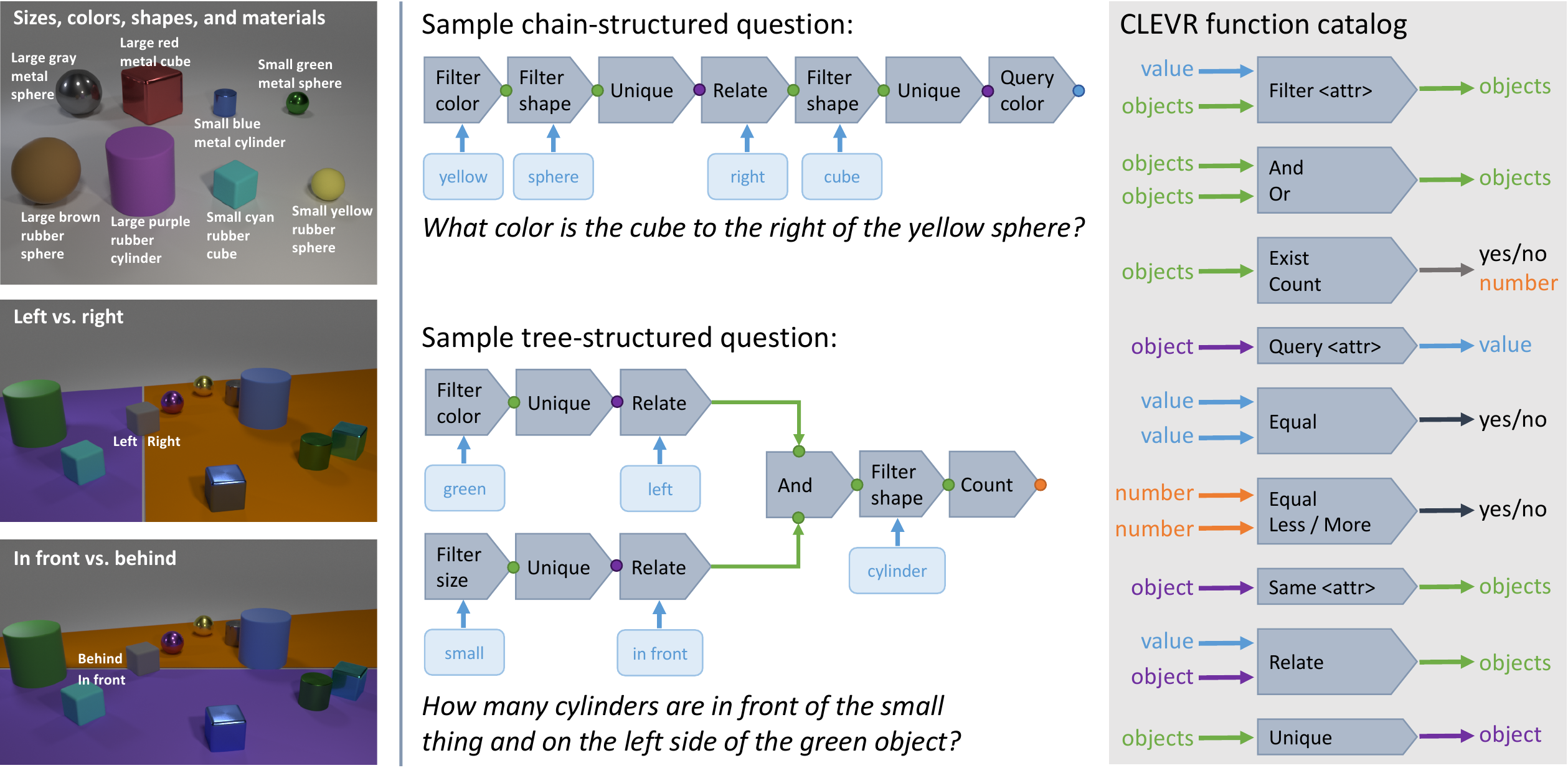}
  \caption{A field guide to the CLEVR universe. \textbf{Left:} Shapes, attributes, and spatial relationships. \textbf{Center:} Examples of questions and their associated functional programs. \textbf{Right:} Catalog of basic functions used to build questions. See Section~\ref{clevr} for details.}
  \vspace{-3mm}
  \label{fig:field-guide}
\end{figure*}

\section{The CLEVR Diagnostic Dataset}
\label{clevr}

CLEVR provides a dataset that requires complex reasoning to solve and that can be used to conduct rich diagnostics to better understand the visual reasoning capabilities of VQA systems.
This requires tight control over the dataset, which we achieve by using synthetic images and automatically generated questions.
The images have associated ground-truth object locations and attributes, and the questions have an associated machine-readable form.
These ground-truth structures allow us to analyze models based on, for example: question type, question topology (chain \vs tree), question length, and various forms of relationships between objects.
Figure~\ref{fig:field-guide} gives a brief overview of the main components of CLEVR, which we describe in detail below.

\paragraph{Objects and relationships.}
The CLEVR universe contains three object shapes (cube, sphere, and cylinder) that come in two absolute sizes (small and large), two materials (shiny ``metal'' and matte ``rubber''), and eight colors.
Objects are spatially related via four relationships: ``left'', ``right'', ``behind'', and ``in front''.
The semantics of these prepositions are complex and depend not only on relative object positions but also on camera viewpoint and context. We found that generating questions that invoke spatial relationships with semantic accord was difficult.
Instead we rely on a simple and unambiguous definition:
projecting the camera viewpoint vector onto the ground plane defines the ``behind'' vector, and one object is behind another if its ground-plane position is further along the ``behind'' vector. The other relationships are similarly defined.
Figure~\ref{fig:field-guide} (left) illustrates the objects, attributes, and spatial relationships in CLEVR. 
The CLEVR universe also includes one non-spatial relationship type that we refer to as the \emph{same-attribute relation}.
Two objects are in this relationship if they have equal attribute values for a specified attribute.

\paragraph{Scene representation.}
Scenes are represented as collections of objects annotated with shape,
size, color, material, and position on the ground-plane. A scene can also be
represented by a \emph{scene graph}~\cite{johnson15,krishna16}, where nodes are
objects annotated with attributes and edges connect spatially related objects.
A scene graph contains all ground-truth information for an image and could be used to replace the vision component of a VQA system with \emph{perfect sight}.

\paragraph{Image generation.}
CLEVR images are generated by randomly sampling a scene graph and rendering it using Blender~\cite{Blender}.
Every scene contains between three and ten objects with random shapes, sizes, materials, colors, and positions.
When placing objects we ensure that no objects intersect, that all objects are at least partially visible,
and that there are small horizontal and vertical margins between the image-plane centers of each pair of objects;
this helps reduce ambiguity in spatial relationships. In each image the positions of the lights and
camera are randomly jittered.

\paragraph{Question representation.}
Each question in CLEVR is associated with a \emph{functional program} that can be \emph{executed} on an image's scene graph, yielding the answer to the question. Functional programs are built from simple basic functions that correspond to elementary operations of visual reasoning such as \emph{querying} object attributes, \emph{counting} sets of objects, or \emph{comparing} values. As shown in Figure~\ref{fig:field-guide}, complex questions can be represented by compositions of these simple building blocks. Full details about each basic function can be found in the supplementary material.

As we will see in Section~\ref{Experiments}, representing questions as functional programs enables rich analysis that would be impossible with natural-language questions. A question's functional program tells us exactly which reasoning abilities are required to solve it, allowing us to compare performance on questions requiring different types of reasoning.

\begin{figure}
  \centering
  \scalebox{0.9}{
    \setlength\tabcolsep{3pt}
    \begin{tabular}{r||cccc}
      & & & Unique & Overlap \\
      Split & Images & Questions & questions & with train \\
      \hline\hline
      Total & 100,000 & 999,968 & 853,554 & - \\
      \hline
      Train & 70,000 & 699,989 & 608,607 & - \\
      Val & 15,000 & 149,991  & 140,448 & 17,338 \\
      Test & 15,000 & 149,988 & 140,352 & 17,335
    \end{tabular}
  }
  \raisebox{2mm}{
    \includegraphics[width=0.245\textwidth]{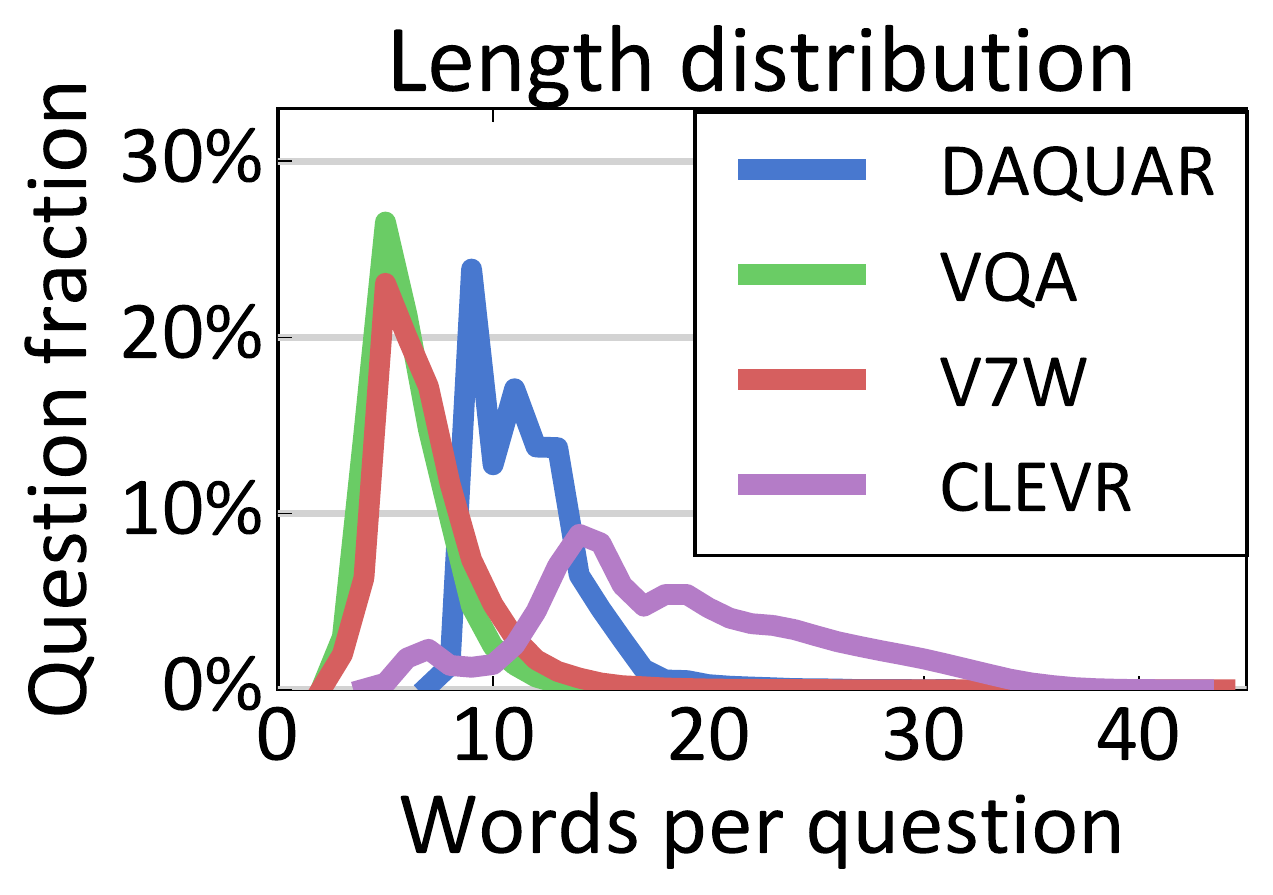}
    \hspace{-2mm}
  }%
  \includegraphics[width=0.25\textwidth]{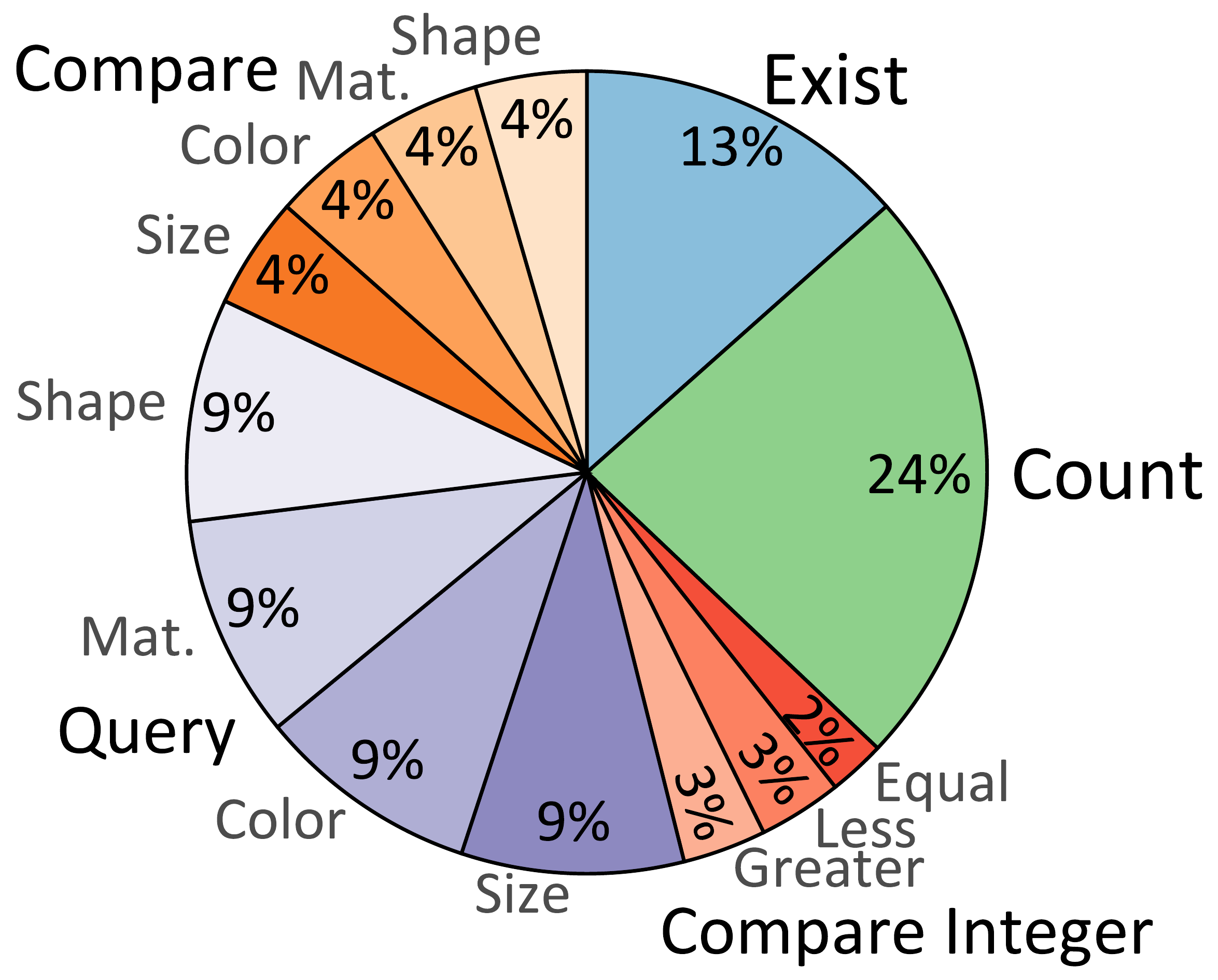}
  \caption{
    \textbf{Top:} Statistics for CLEVR; the majority of questions
    are unique and few questions from the val and test sets appear in the training set.
    \textbf{Bottom left:} Comparison of question lengths for different VQA
    datasets; CLEVR questions are generally much longer. \textbf{Bottom right:}
    Distribution of question types in CLEVR.
  }
  \vspace{-4mm}
  \label{fig:stats}
\end{figure}

We categorize questions by \emph{question type}, defined by the outermost function in the question's program; for example the questions in Figure~\ref{fig:field-guide} have types \emph{query-color} and \emph{exist}. Figure~\ref{fig:stats} shows the number of questions of each type.

\paragraph{Question families.}
We must overcome several key challenges to generate a VQA dataset using functional programs. Functional building blocks can be used to construct an infinite number of possible functional programs, and we must decide which program structures to consider. We also need a method for converting functional programs to natural language in a way that minimizes question-conditional bias. We solve these problems using \emph{question families}.

A question family contains a template for constructing functional programs and several text templates providing multiple ways of expressing these programs in natural language. For example, the question \emph{``How many red things are there?''} can be formed by instantiating the text template \emph{``How many $<$C$>$ $<$M$>$ things are there?''}, binding the parameters \emph{$<$C$>$} and \emph{$<$M$>$} (with types ``color'' and ``material'') to the values \emph{red} and \texttt{nil}. The functional program \emph{count(filter\_color(red, scene()))} for this question
can be formed by instantiating the associated program template
\setlength{\abovedisplayskip}{2pt}
\setlength{\belowdisplayskip}{2pt}
\begin{flalign}
  \emph{count(filter\_color($<$C$>$,~filter\_material($<$M$>$,~scene())))}\nonumber
\end{flalign}
with the same values, using the convention that functions taking a \texttt{nil} input are removed after instantiation.

CLEVR contains a total of 90 question families, each with a single program template and an average of four text templates. Text templates were generated by manually writing one or two templates per family and then crowdsourcing question rewrites. To further increase language diversity we use a set of synonyms for each shape, color, and material. With up to 19 parameters per template, a small number of families can generate a huge number of unique questions; Figure~\ref{fig:stats} shows that of the nearly one million questions in CLEVR, more than 853k are unique.
CLEVR can easily be extended by adding new question families.

\paragraph{Question generation.}
\label{sec:question-gen}
Generating a question for an image is conceptually simple: we choose a question family, select values for each of its template parameters,
execute the resulting program on the image's scene graph to find the answer, and use one of the text templates from the question family
to generate the final natural-language question.

However, many combinations of values give rise to questions which are either \emph{ill-posed} or \emph{degenerate}. 
The question \emph{``What color is the cube to the right of the sphere?"} would be \emph{ill-posed} if there were many cubes right of the sphere,
or \emph{degenerate} if there were only one cube in the scene since the reference to the sphere would then be unnecessary. Avoiding such ill-posed and degenerate questions is critical to ensure the correctness and complexity of our questions.

A na\"{i}ve solution is to randomly sample combinations of values and reject those which lead to ill-posed or degenerate questions.
However, the number of possible configurations for a question family is exponential in its number of parameters, and most of them are undesirable.
This makes brute-force search intractable for our complex question families. 

Instead, we employ a depth-first search to find valid values for instantiating question families. At each step of the search, we use ground-truth scene information
to prune large swaths of the search space which are guaranteed to produce undesirable questions; for example we need not entertain questions
of the form \emph{``What color is the $<$S$>$ to the $<$R$>$ of the sphere''} for scenes that do not contain spheres.

Finally, we use rejection sampling to produce an approximately uniform answer distribution for each question family;
this helps minimize question-conditional bias since all questions from the same family share linguistic structure.

\section{VQA Systems on CLEVR}
\label{Experiments}

\begin{figure*}[!ht]
  \centering
  \includegraphics[height=0.132\textwidth,trim={5px 2px 0 0},clip]{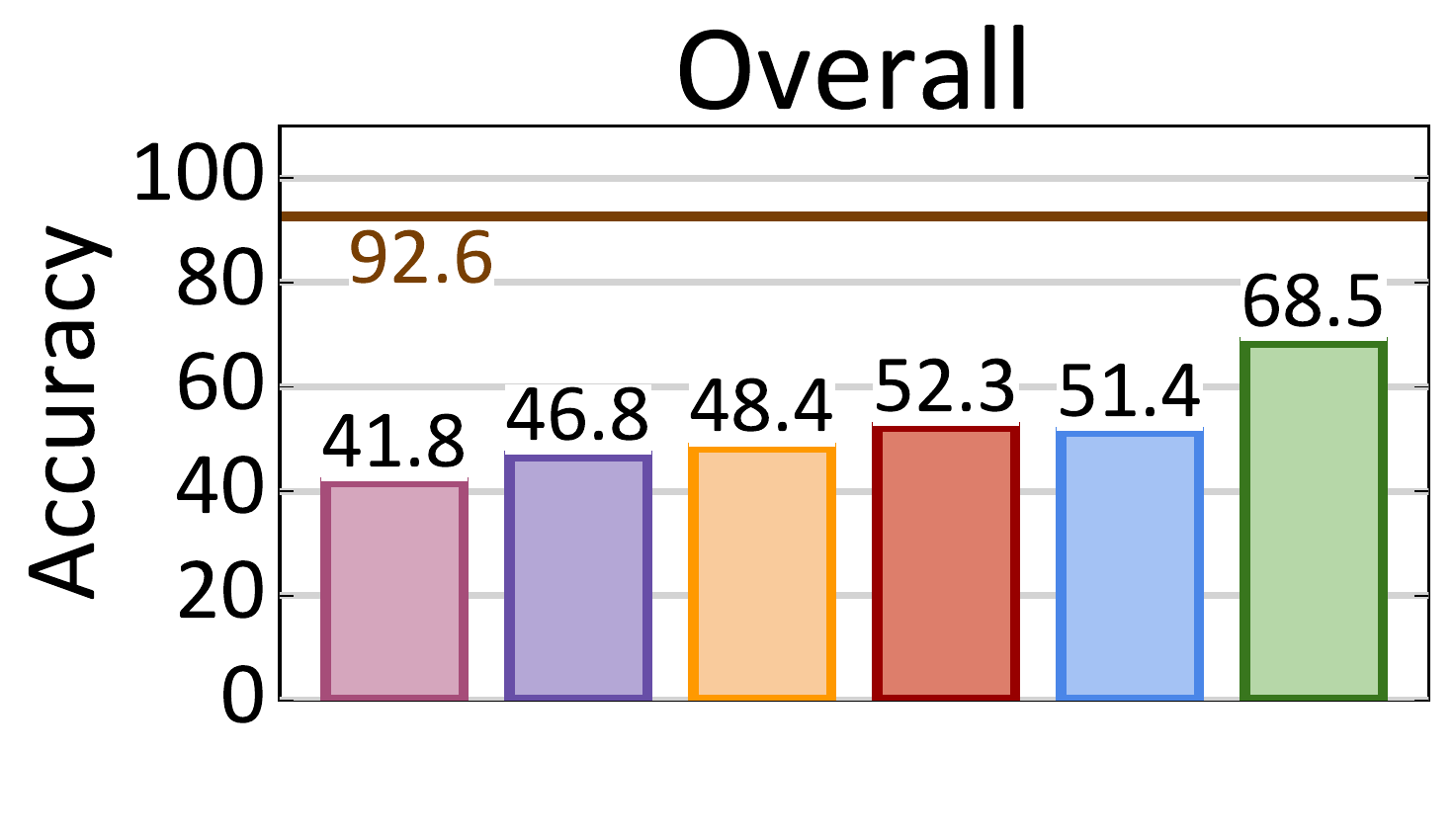}
  \hspace{-4px}
  \includegraphics[height=0.132\textwidth,trim={50px 2px 0 0},clip]{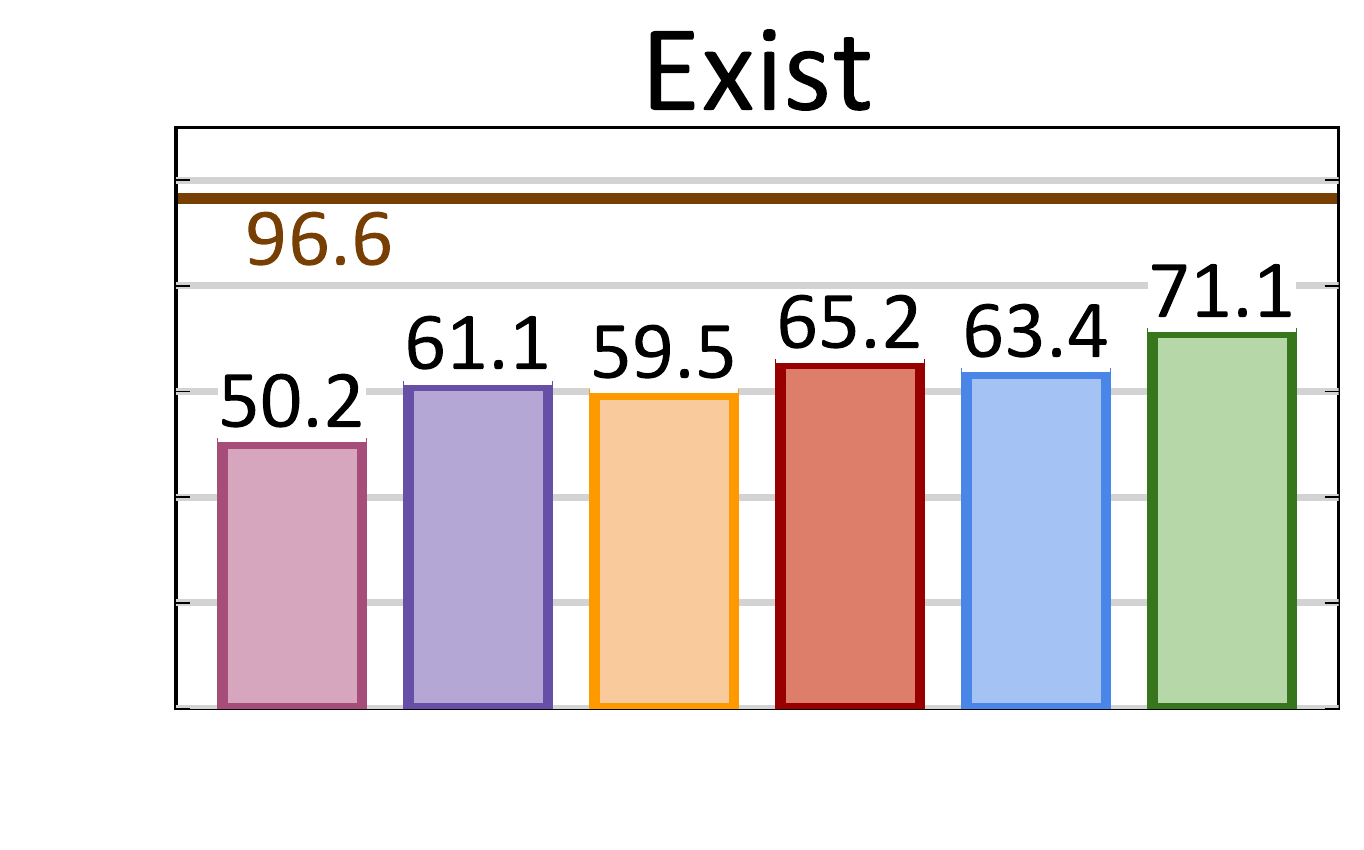}
  \hspace{-4px}
  \includegraphics[height=0.132\textwidth,trim={50px 2px 0 0},clip]{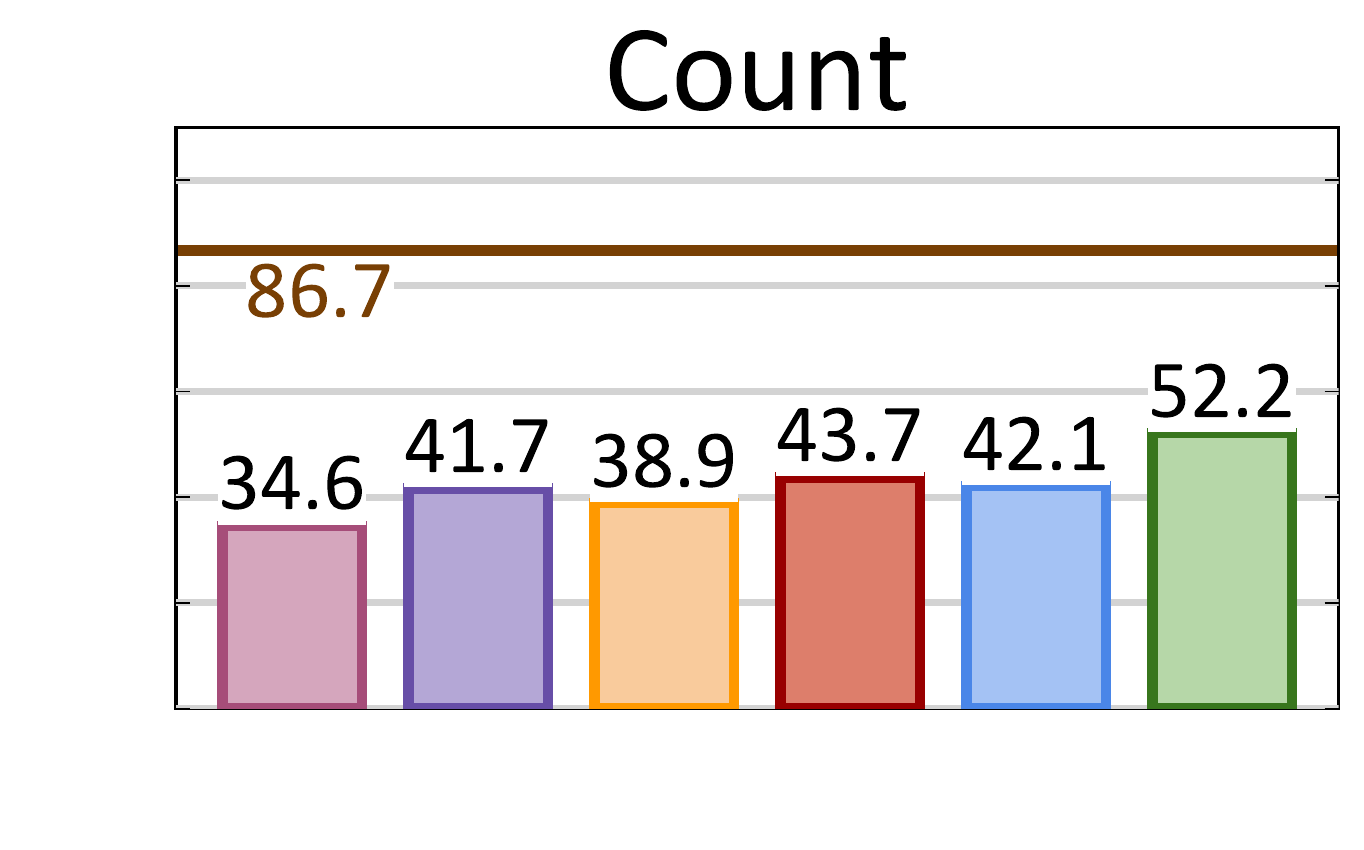}
  \hspace{-4px}
  \includegraphics[height=0.135\textwidth,trim={50px 2px 0 0},clip]{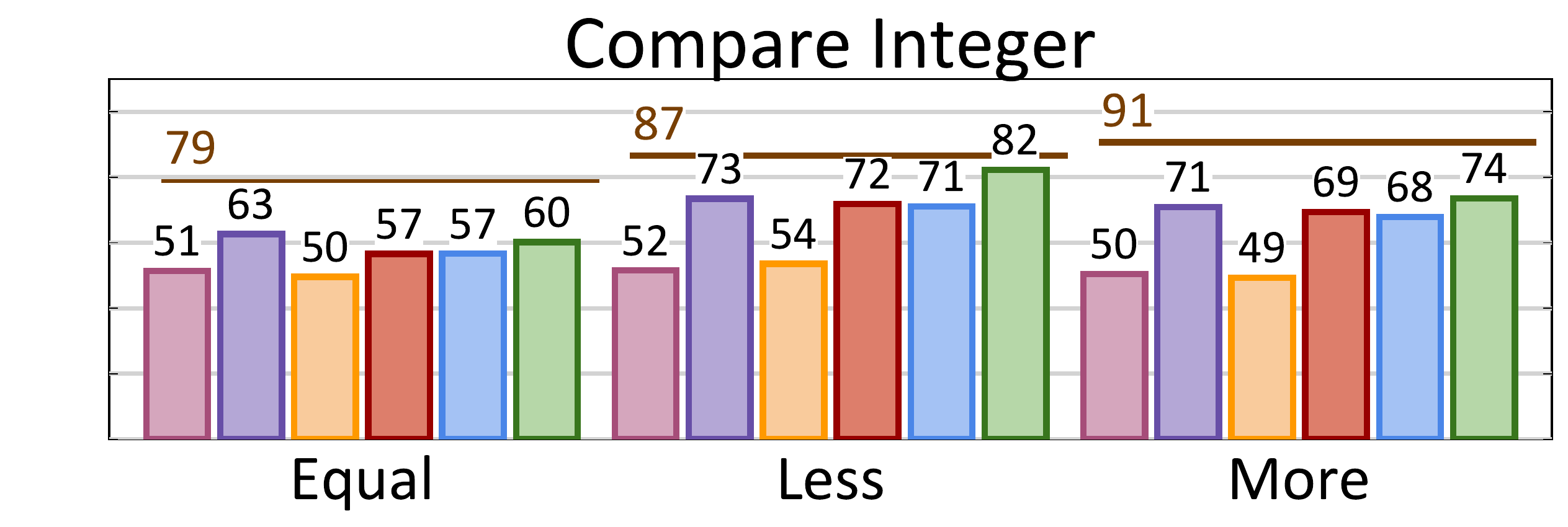} \\*[-1mm]
  \includegraphics[height=0.132\textwidth,trim={0 8px 0 8px}]{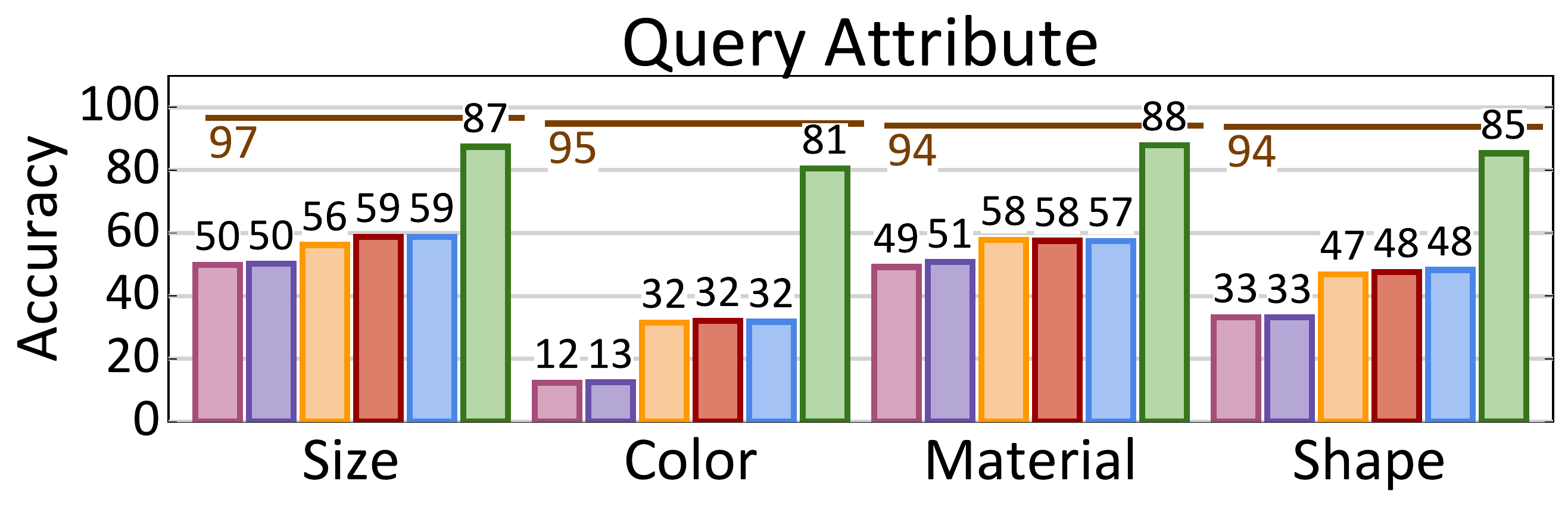}
  \includegraphics[height=0.132\textwidth,trim={50px 8px 0 8px},clip]{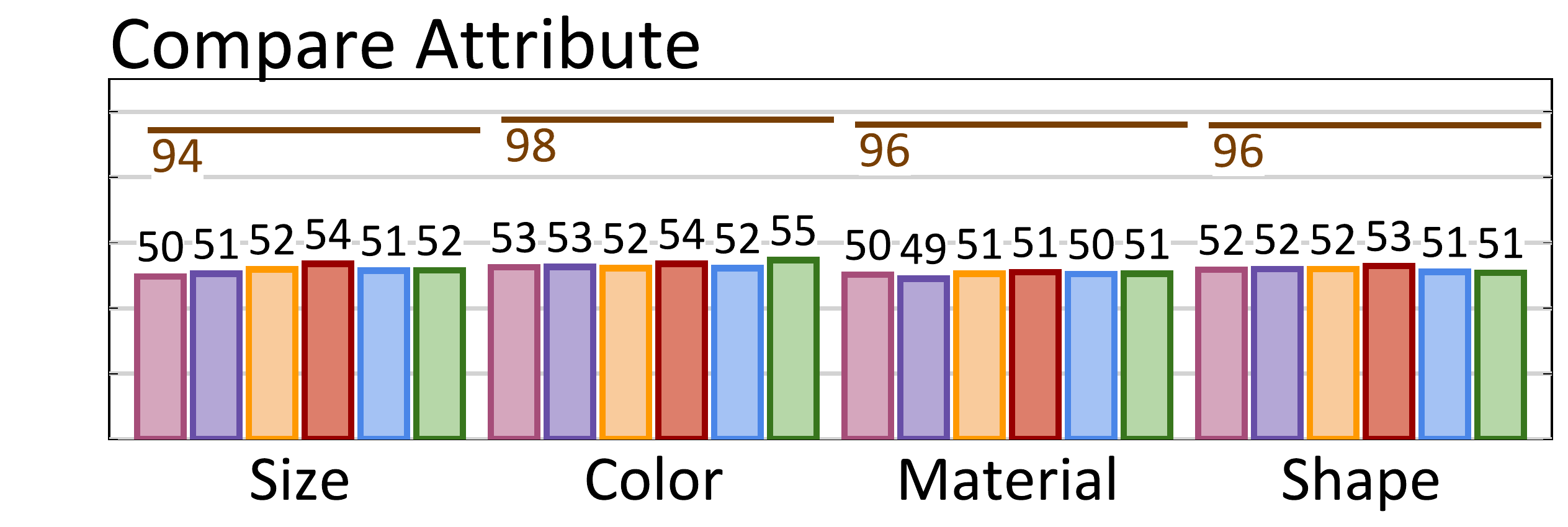}
  \raisebox{2px}{\includegraphics[height=0.13\textwidth]{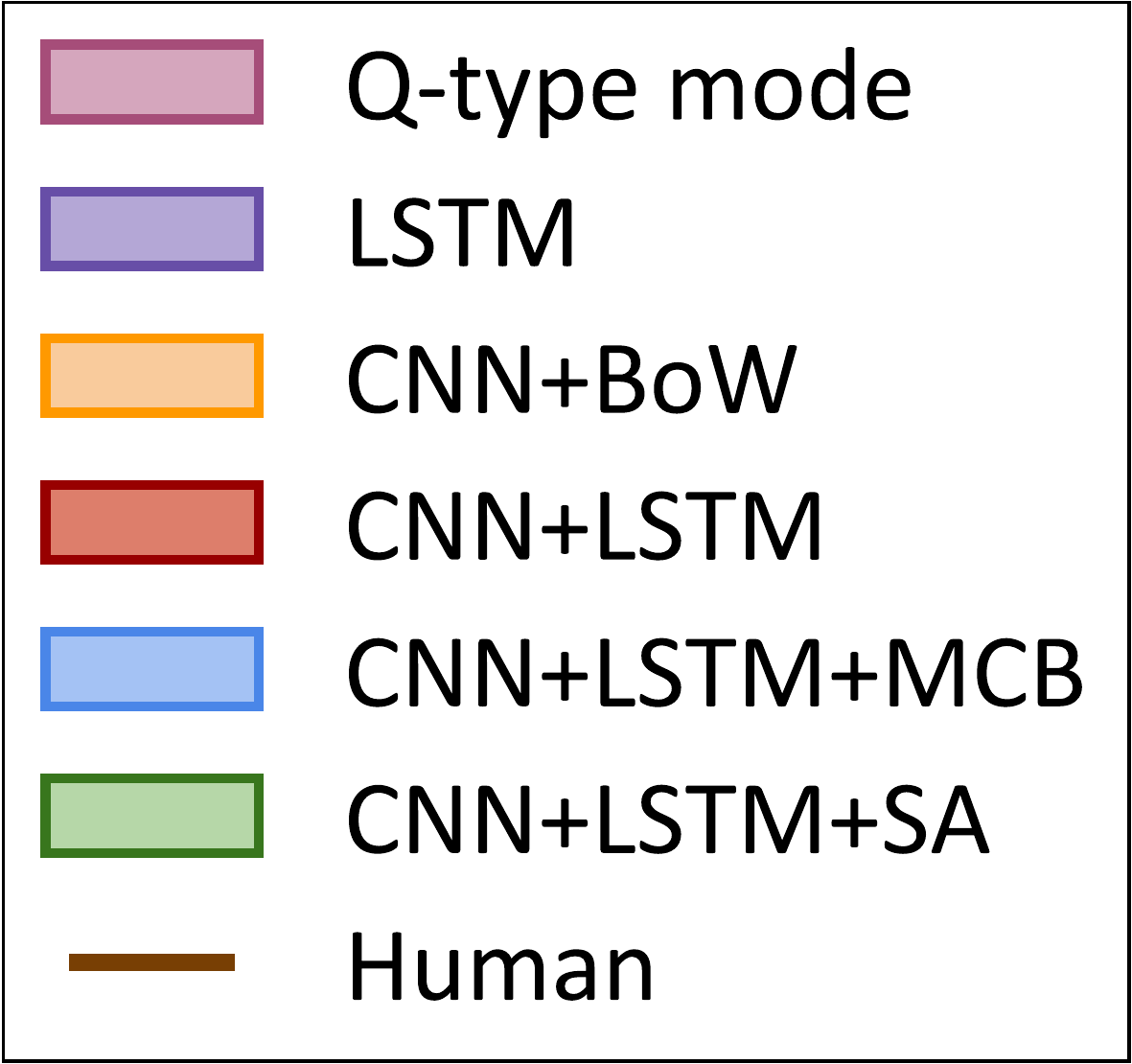}}
  \caption{
    Accuracy per question type of the six VQA methods on the CLEVR dataset (higher is better). Figure best viewed in color.
  }
  \vspace{-4mm}
  \label{fig:acc-type}
\end{figure*}

\subsection{Models}
VQA models typically represent images with features from pretrained CNNs and
use word embeddings or recurrent networks to represent questions and/or answers.
Models may train recurrent networks for answer generation~\cite{gao15,malinowski15,wu16},
multiclass classifiers over common answers~\cite{antol15,hiatt16,ma15,ren15,zhou15,zhu15},
or binary classifiers on image-question-answer triples~\cite{fukui16,jabri16,shih16}.
Many methods incorporate attention over the image~\cite{fukui16,shih16,yang16,zhu15,xu16}
or question~\cite{hiatt16}. Some methods incorporate memory~\cite{xiong2016dynamic} or
dynamic network architectures~\cite{andreas16b,andreas16}.

Experimenting with all methods is logistically challenging, so we reproduced a
representative subset of methods: baselines that do not look at the image
(Q-type mode, LSTM), a simple baseline (CNN+BoW) that performs near
state-of-the-art~\cite{jabri16,zhou15}, and more sophisticated methods using
recurrent networks (CNN+LSTM), sophisticated feature pooling
(CNN+LSTM+MCB), and spatial attention (CNN+LSTM+SA).\footnote{We performed initial
experiments with dynamic module networks~\cite{andreas16b} but its parsing heuristics
did not generalize to the complex questions in CLEVR so it did not work out-of-the-box;
see supplementary material.} These are described in detail below.

\textbf{Q-type mode:} Similar to the ``per Q-type prior"
method in \cite{antol15}, this baseline predicts the most frequent
training-set answer for each question's type.

\textbf{LSTM:} Similar to ``LSTM Q'' in \cite{antol15},
the question is processed with learned word embeddings followed by a
word-level LSTM~\cite{hochreiter97}. The final LSTM hidden state is passed
to a multi-layer perceptron (MLP) that predicts a distribution over answers.
This method uses no image information so it can only model
question-conditional bias.

\textbf{CNN+BoW:} Following \cite{zhou15}, the
question is encoded by averaging word vectors for each word in the question and 
the image is encoded using features from a convolutional network (CNN). The question
and image features are concatenated and passed to a MLP which
predicts a distribution over answers. We use word vectors trained on the GoogleNews
corpus~\cite{mikolov13}; these are not fine-tuned during training.

\textbf{CNN+LSTM:}
As above, images and questions are encoded using CNN features and final LSTM hidden states, respectively.
These features are concatenated and passed to an MLP that predicts an answer distribution.

\textbf{CNN+LSTM+MCB:} Images and questions are encoded as above,
but instead of concatenation, their features are pooled using compact multimodal pooling (MCB)~\cite{fukui16,gao2016compact}.

\textbf{CNN+LSTM+SA:} Again, the question and image are encoded using a CNN and LSTM, respectively.
Following~\cite{yang16}, these representations are combined using one or more rounds of soft spatial
attention and the final answer distribution is predicted with an MLP.

\textbf{Human:} We used Mechanical Turk to collect human responses for 5500 random questions from the
test set, taking a majority vote among three workers for each question.

\paragraph{Implementation details.}
Our CNNs are ResNet-101 models pretrained on ImageNet~\cite{he16} that are not finetuned; images are resized to $224 \! \times \! 224$ prior to 
feature extraction. CNN+LSTM+SA extracts features from
the last layer of the \verb|conv4| stage, giving $14\times14\times1024$-dimensional features.
All other methods extract features from the final average pooling layer, giving
2048-dimensional features.
LSTMs use one or two layers with 512 or 1024 units per layer.
MLPs use ReLU functions and dropout~\cite{srivastava14}; they have one or two hidden
layers with between 1024 and 8192 units per layer. All models are trained using Adam~\cite{kingma14}.

\paragraph{Experimental protocol.}
CLEVR is split into train, validation, and test sets (see Figure~\ref{fig:stats}).
We tuned hyperparameters (learning rate, dropout, word vector size, number and size
of LSTM and MLP layers) independently per model based on the validation error.
All experiments were designed on the validation set; after finalizing the
design we ran each model once on the test set.
\emph{All experimental findings generalized from the validation set to the test set.}

\subsection{Analysis by Question Type}
\label{sec:question-type}
We can use the program representation of questions to analyze model
performance on different forms of reasoning. We first evaluate performance on each question type,
defined as the outermost function in the program.
Figure~\ref{fig:acc-type} shows results and detailed findings are discussed below.

\textbf{Querying attributes:}
Query questions ask about an attribute of a particular object
(\eg \textit{``What color is the thing right of the red sphere?''}).
The CLEVR world has two sizes, eight colors, two materials, and three shapes.
On questions asking about these different attributes, Q-type mode and LSTM obtain accuracies close
to 50\%, 12.5\%, 50\%, and 33.3\% respectively, showing that the dataset has minimal question-conditional
bias for these questions.
CNN+LSTM+SA substantially outperforms all other models on these questions;
its attention mechanism may help it focus on the target object and identify its attributes.

\textbf{Comparing attributes:}
Attribute comparison questions ask whether two objects have the same value for some attribute
(\eg \textit{``Is the cube the same size as the sphere?''}).
The only valid answers are ``yes'' and ``no''.
Q-Type mode and LSTM achieve accuracies close to 50\%, confirming there is no dataset bias for these questions. Unlike attribute-query questions, attribute-comparison questions require a limited form of memory: 
models must identify the attributes of two objects and keep them in memory to compare them.
Interestingly, none of the models are able to do so: all models have an accuracy of approximately 50\%.
This is also true for the CNN+LSTM+SA model, suggesting that its attention mechanism is not capable of attending to two objects at once to compare them.
This illustrates how CLEVR can reveal limitations of models and motivate follow-up research, \eg, augmenting attention models with explicit memory.

\begin{figure}[!tb]
  \centering
  \includegraphics[height=0.09\textwidth,trim={2px 0 0 0},clip]{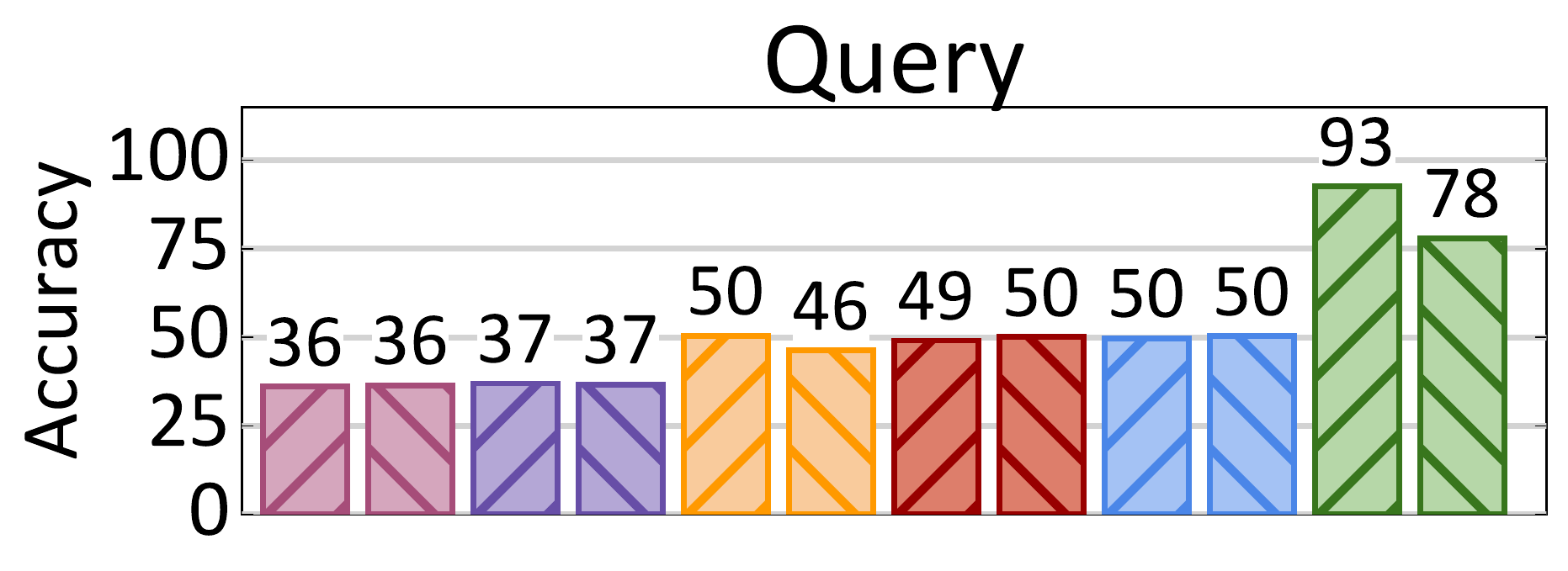}%
  \raisebox{7mm}{
    \begin{minipage}{0.23\textwidth}
      \centering
      \includegraphics[width=0.9\textwidth]{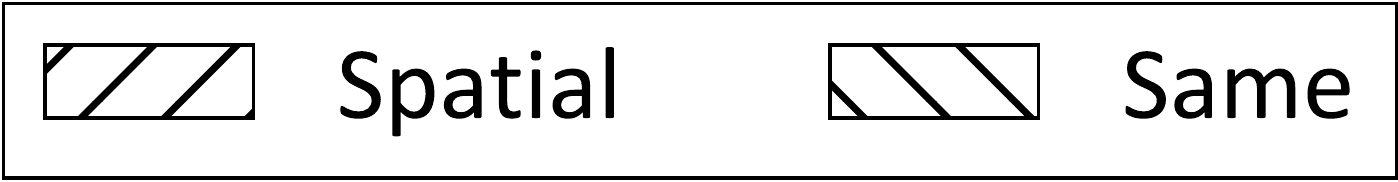}
      \begin{minipage}{0.41\textwidth}
        \scriptsize
        \textit{What color is the cube to the right of the sphere?}
      \end{minipage}%
      \hspace{0.03\textwidth}%
      \begin{minipage}{0.51\textwidth}
        \scriptsize
        \textit{What color is the cube that is the same size as the sphere?}
      \end{minipage}
    \end{minipage}
  } \\*[-1mm]
  \hspace{-2.5mm}\includegraphics[height=0.09\textwidth,trim={2px 0 0 0},clip]{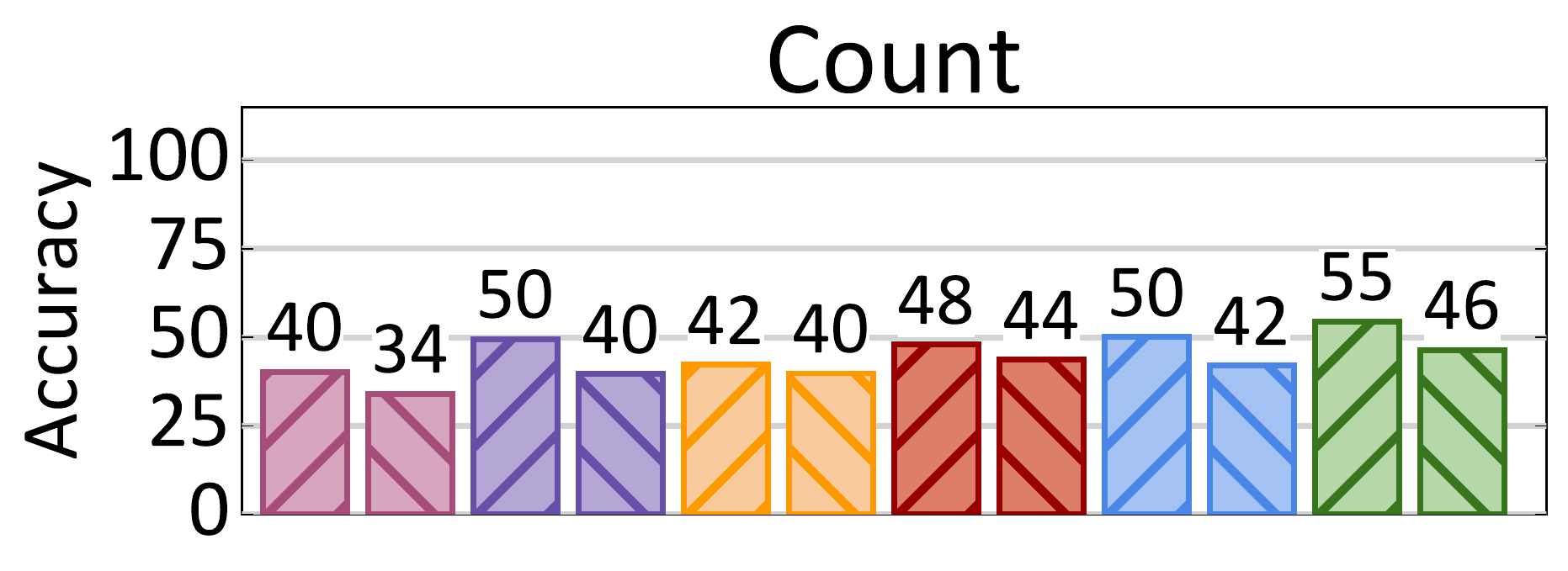}
  \includegraphics[height=0.09\textwidth,trim={80px 0 0 0},clip]{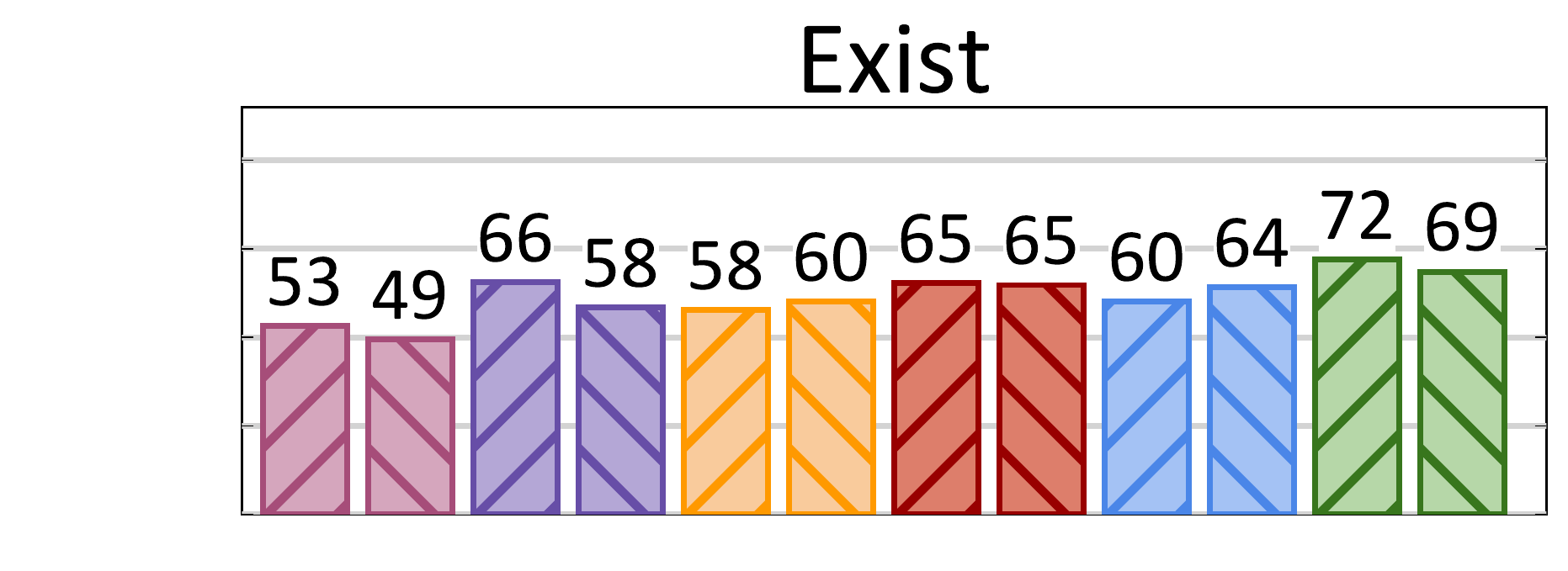} \\*[-0.5mm]
  \hspace{5mm}
  \includegraphics[width=0.35\textwidth]{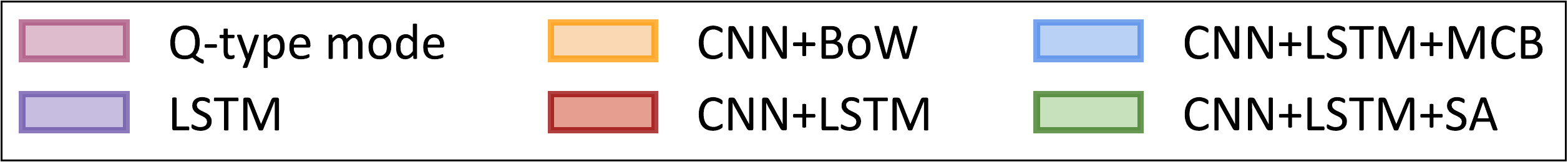}
  \caption{
    Accuracy on questions with a single \emph{spatial relationship} \vs a single
    \emph{same-attribute} relationship. For \emph{query} and \emph{count} questions,
    models generally perform worse on questions with \emph{same-attribute}
    relationships. Results on \emph{exist} questions are mixed.
  }
  \vspace{-3mm}
  \label{fig:same-relate}
\end{figure}

\textbf{Existence:}
Existence questions ask whether a certain type of object is present
(\eg, \emph{``Are there any cubes to the right of the red thing?''}).
The 50\% accuracy of Q-Type mode shows that both answers are \emph{a priori} equally likely, but the LSTM result of 60\% does suggest a question-conditional bias. There may be correlations between question length and answer: questions with more filtering operations (\eg, ``large red cube'' \vs ``red cube'') may be more likely to have ``no'' as the answer. Such biases may be present even with uniform answer distributions per question
family, since questions from the same family may have different numbers of filtering functions.
CNN+LSTM(+SA) outperforms LSTM, but its performance is still quite low.

\textbf{Counting:}
Counting questions ask for the number of objects fulfilling some conditions
(\eg \textit{``How many red cubes are there?''});
valid answers range from zero to ten.
Images have three and ten objects and counting questions refer to subsets of objects, so ensuring a
uniform answer distribution is very challenging; our rejection sampler therefore
pushes towards a uniform distribution for these questions rather than enforcing it as a hard constraint. This results in
a question-conditional bias, reflected in the 35\% and 42\% accuracies achieved by Q-type mode and LSTM.
CNN+LSTM(+MCB) performs on par with LSTM, suggesting that 
CNN features contain little information relevant to counting. CNN+LSTM+SA performs slightly better, but at 52\% its
absolute performance is low.

\begin{figure}[!ht]
  \centering
  \includegraphics[height=0.0941\textwidth,trim={5px 0 0 0},clip]{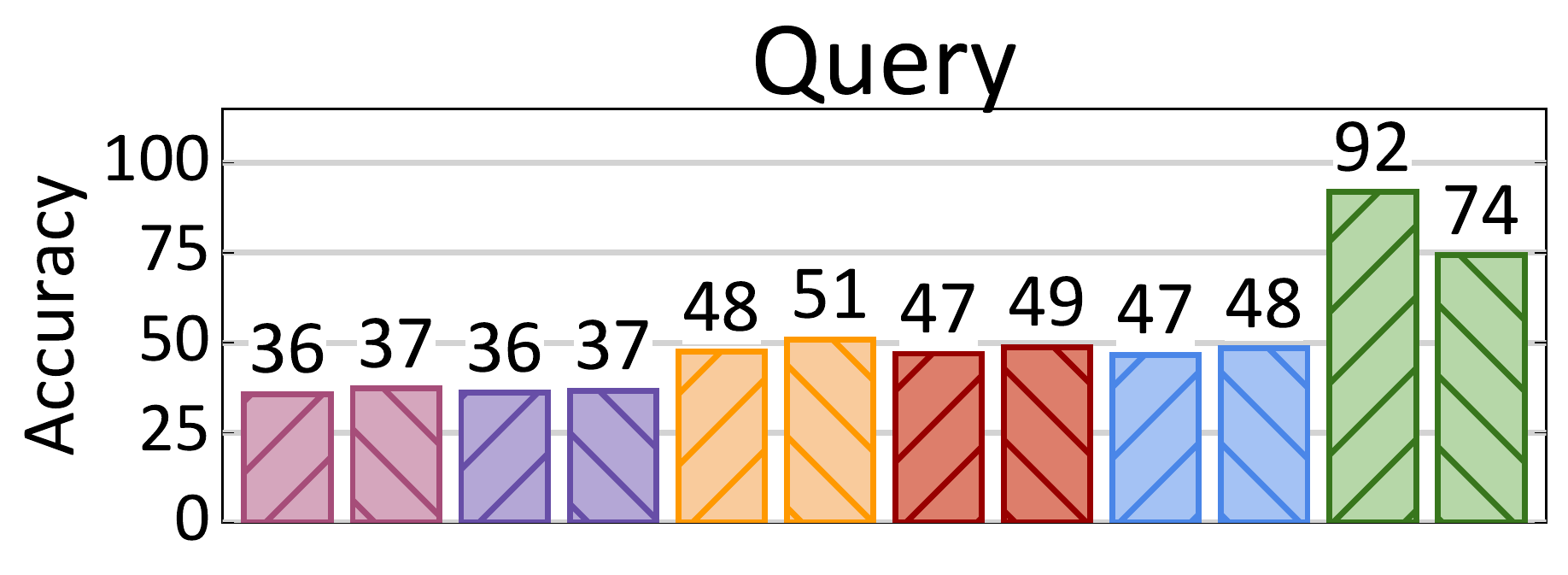}%
  \includegraphics[height=0.0941\textwidth,trim={70px 0 0 0},clip]{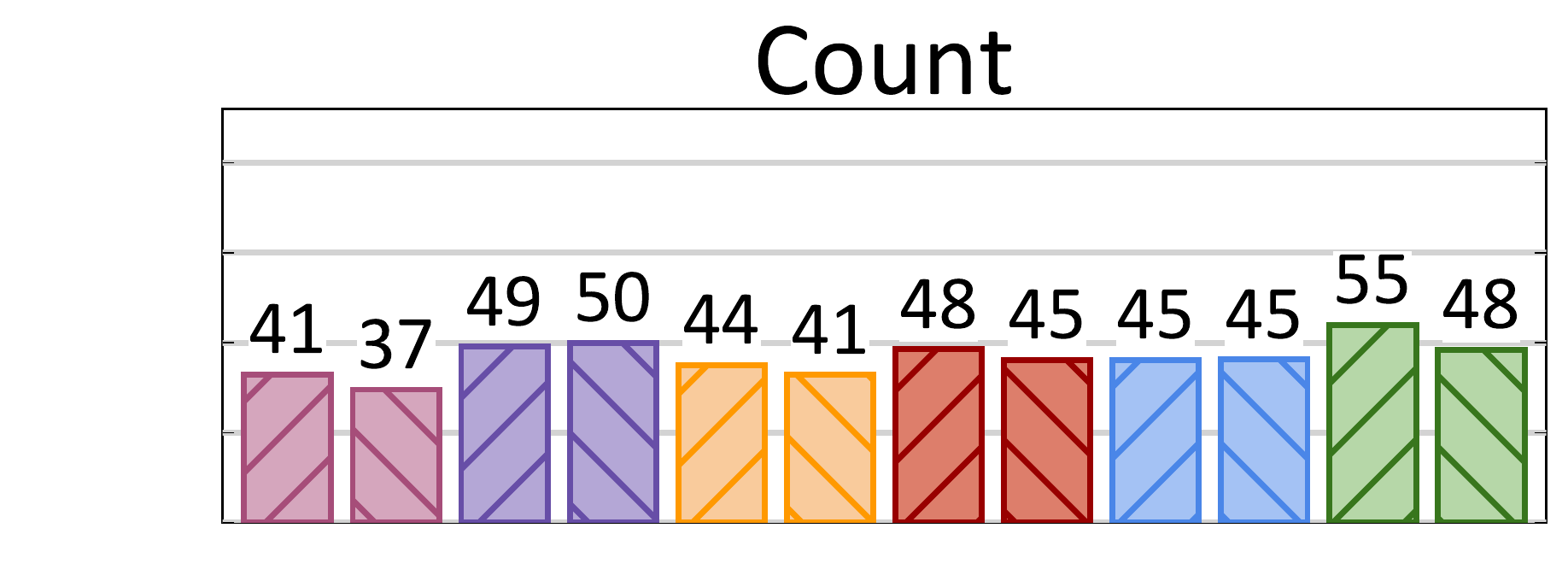}
  \raisebox{13.5mm}{
    \begin{minipage}{0.3\textwidth}
      \centering
      \includegraphics[width=0.8\textwidth]{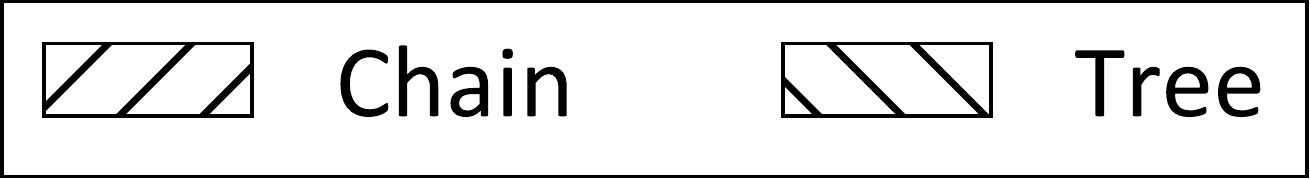}
      \begin{minipage}{0.4\textwidth}
        \scriptsize
        \textit{How many cubes are to the right of the sphere that is to the left of the red thing?}
      \end{minipage}%
      \hspace{0.07\textwidth}
      \begin{minipage}{0.38\textwidth}
        \scriptsize
        \textit{How many cubes are both to the right of the sphere and left of the red thing?}
      \end{minipage}
    \end{minipage}
  }
  \raisebox{5mm}{
    \includegraphics[width=0.13\textwidth]{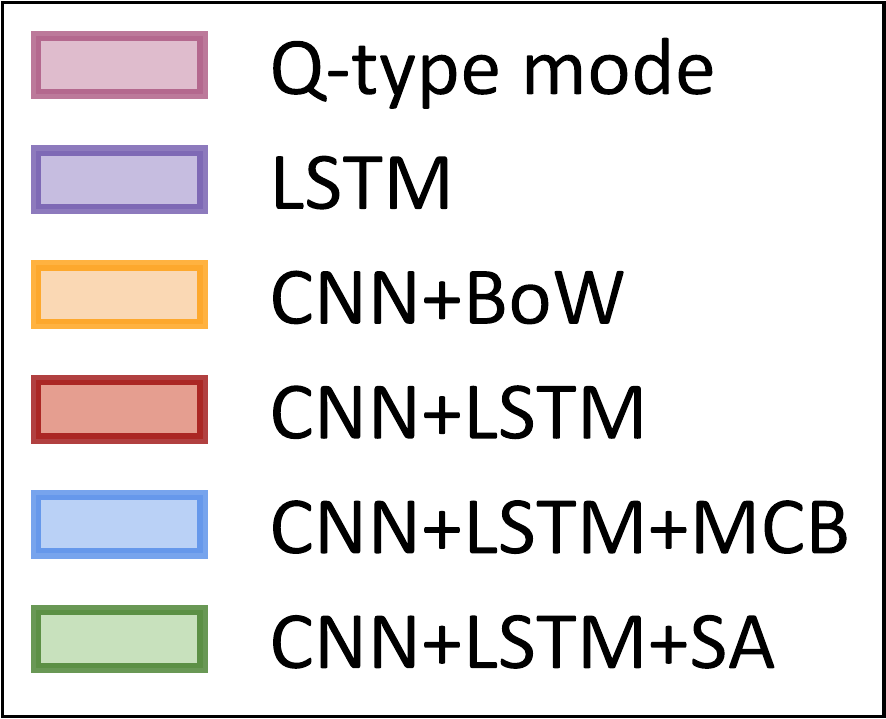}
  }
  \vspace{-5mm}
  \caption{
    Accuracy on questions with two spatial relationships,
    broken down by question topology: chain-structured questions \vs tree-structured
    questions joined with a logical AND operator.
  }
  \label{fig:and-or}
  \vspace{-2mm}
\end{figure}

\textbf{Integer comparison:}
Integer comparison questions ask which of two object sets is larger
(\eg \textit{``Are there fewer cubes than red things?''});
this requires counting, memory, and comparing integer quantities.
The answer distribution is unbiased (see Q-Type mode) but a set's size may correlate with the length of its description, explaining
the gap between LSTM and Q-type mode. CNN+BoW performs no better than chance: BoW mixes the words describing each set, making it impossible for the learner to discriminate between them. CNN+LSTM+SA outperforms LSTM on ``less'' and ``more'' questions, but no model outperforms LSTM on ``equal'' questions. Most models perform better on ``less'' than ``more'' due to asymmetric question families.

\subsection{Analysis by Relationship Type}
\label{sec:relationship-type}
CLEVR questions contain two types of relationships: \emph{spatial} and \emph{same-attribute} (see Section~\ref{clevr}).
We can compare the relative difficulty of these two types by comparing model performance on questions with a single
spatial relationship and questions with a single same-attribute relationship; results are shown in
Figure~\ref{fig:same-relate}. On query-attribute and counting questions we see that same-attribute questions are
generally more difficult; the gap between CNN+LSTM+SA on spatial and same-relate query questions is particularly large
(93\% \vs 78\%). Same-attribute relationships may require a model to keep attributes of one object ``in memory''
for comparison, suggesting again that models augmented with explicit memory may perform better on these questions.

\subsection{Analysis by Question Topology}
\label{sec:topology}
We next evaluate model performance on different question topologies: \emph{chain-structured} questions
\vs \emph{tree-structured} questions with two branches joined by a logical AND (see Figure~\ref{fig:field-guide}).
In Figure~\ref{fig:and-or}, we compare performance on chain-structured questions with two spatial relationships
vs. tree-structured questions with one relationship along each branch. On query questions, CNN+LSTM+SA shows
a large gap between chain and tree questions (92\% \vs 74\%); on count questions, CNN+LSTM+SA slightly outperforms
LSTM on chain questions (55\% \vs 49\%) but no method outperforms LSTM on tree questions. Tree questions may be
more difficult since they require models to perform two subtasks in parallel before fusing their results.

\begin{figure}[!ht]
  \centering
  \raisebox{3mm}{
    \begin{minipage}[b]{0.2\textwidth}
      \scriptsize
      \textbf{Question:}
      \textit{There is a large object that is on the left side of the large blue
        cylinder in front of the rubber cylinder on the right side of the purple
        shiny thing; what is its shape?}
      \\*
      \textbf{Effective Question:}
      \textit{What shape is a large object left of a cylinder?}
    \end{minipage}
  }
  \hspace{3mm}\includegraphics[width=0.19\textwidth]{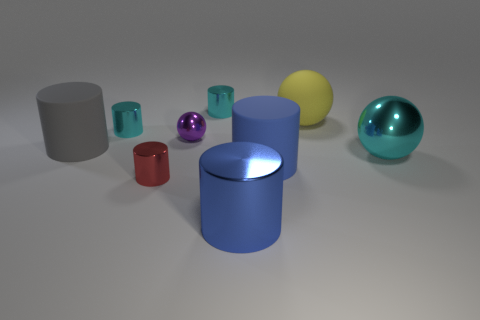}
  \includegraphics[height=0.13\textwidth,trim={5px 0 0 0},clip]{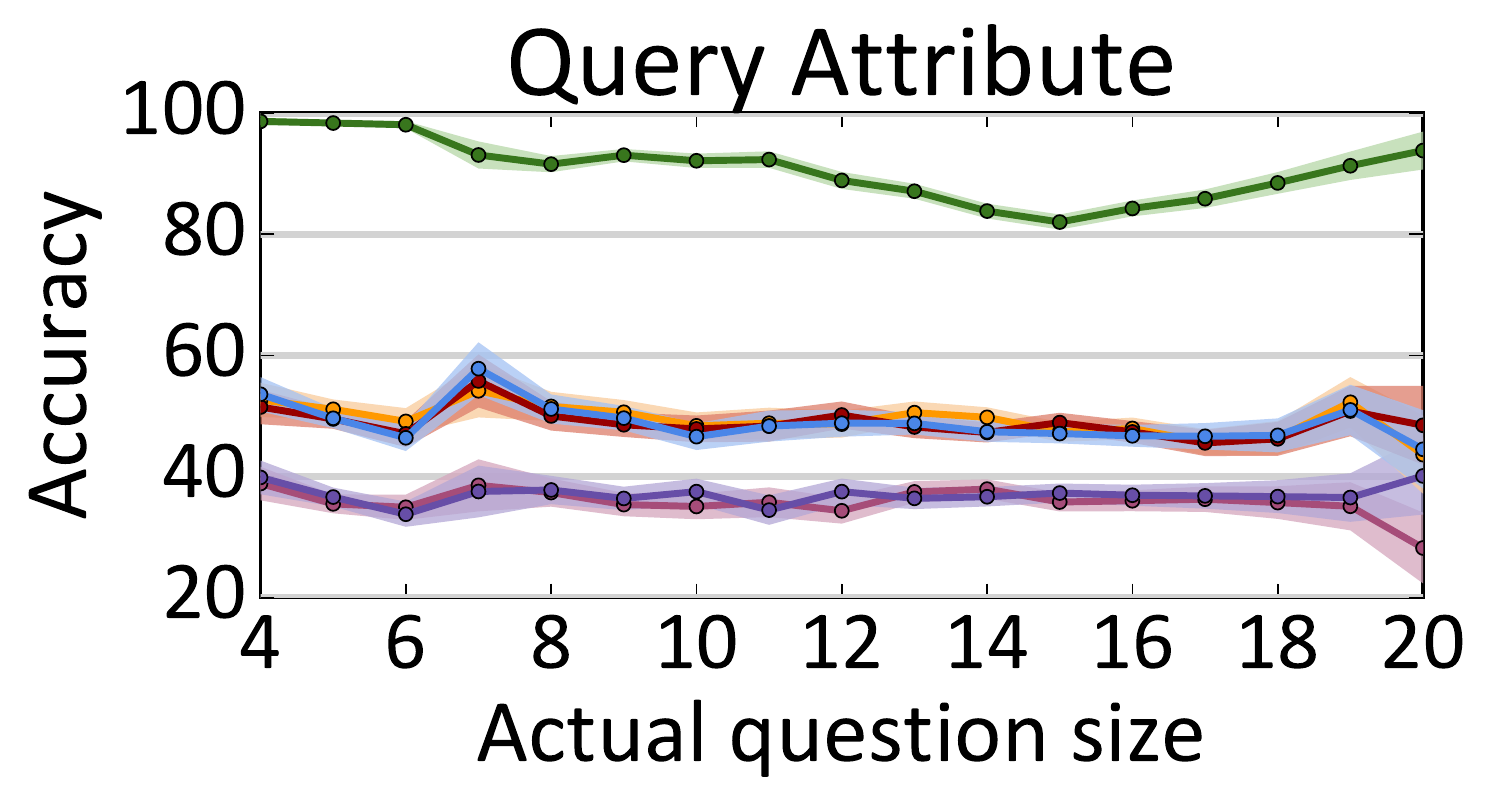}
  \hspace{-5px}
  \includegraphics[height=0.13\textwidth,trim={60px 0 0 0},clip]{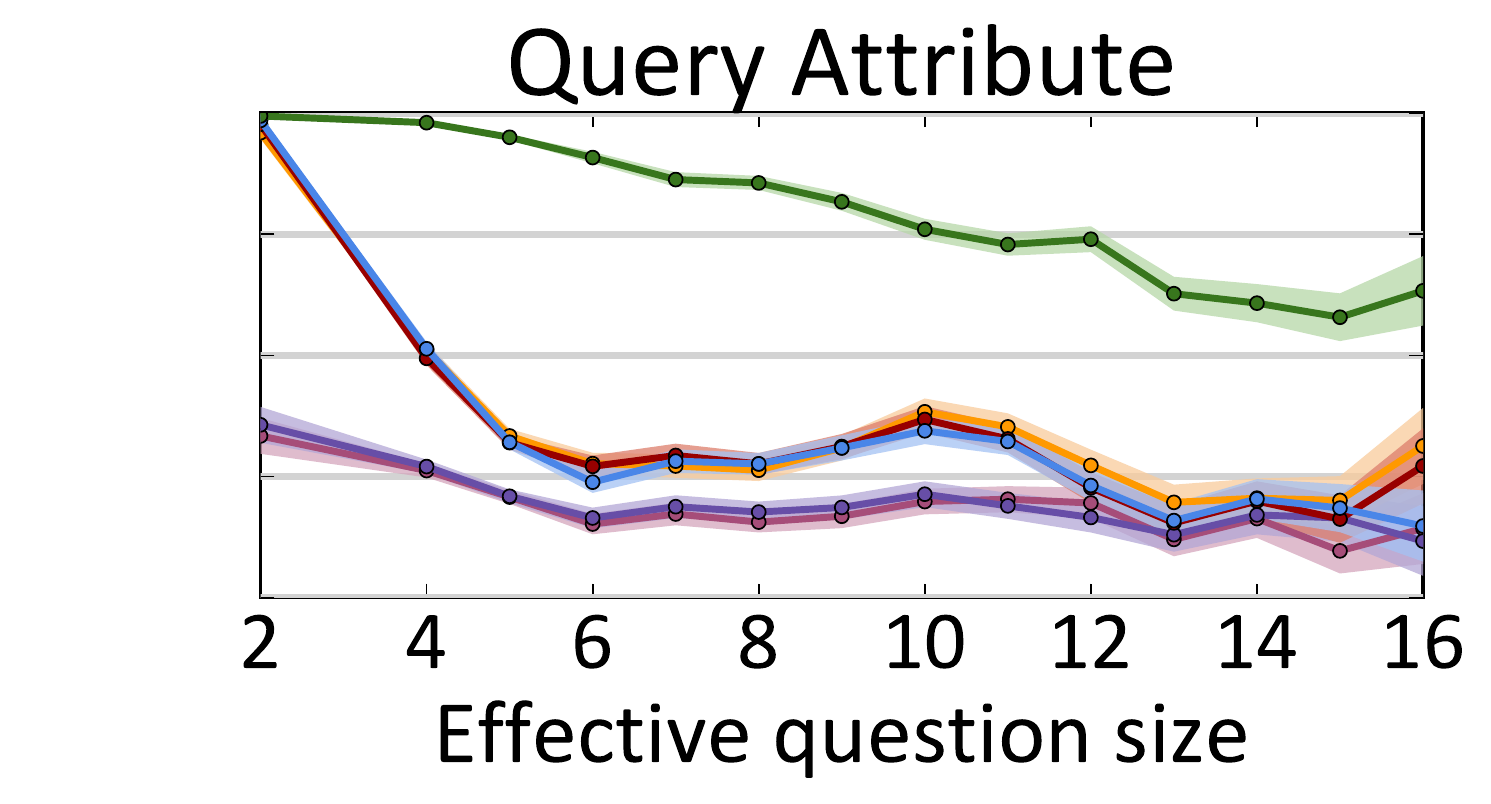} \\*
  \hspace{4mm}
  \includegraphics[width=0.40\textwidth]{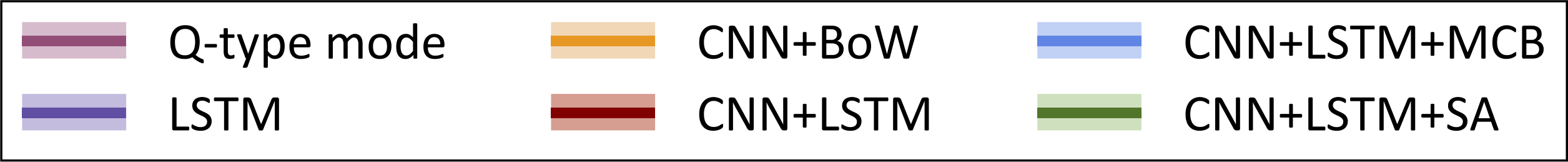}
  \caption{
    \textbf{Top}: Many questions can be answered correctly without
    correctly solving all subtasks. For a given question and scene we can prune
    functions from the question's program to generate an \emph{effective question}
    which is shorter but gives the same answer.\hspace{1cm}
    \textbf{Bottom}: Accuracy on query questions \vs actual and
    effective question size. Accuracy decreases with effective question size but
    not with actual size. Shaded area shows a 95\% confidence interval.
  }
  \vspace{-5mm}
  \label{fig:effective-size}
\end{figure}

\subsection{Effect of Question Size}
\label{sec:effective-size}
Intuitively, longer questions should be harder since they involve more reasoning steps.
We define a question's \emph{size} to be the number of functions in its program, and in
Figure~\ref{fig:effective-size} (bottom left) we show accuracy on query-attribute questions as
a function of question size.\footnote{We exclude questions with same-attribute relations
since their max size is 10, introducing unwanted correlations between size and difficulty.
Excluded questions show the same trends (see supplementary material).}
Surprisingly accuracy appears unrelated to question size.

However, many questions can be correctly answered even when some subtasks are not
solved correctly. For example, the question in Figure~\ref{fig:effective-size} (top)
can be answered correctly without identifying the correct large blue cylinder, because all
large objects left of a cylinder are cylinders.

To quantify this effect, we define the \emph{effective question} of an image-question pair: we prune functions from
the question's program to find the smallest program that, when executed on the scene graph for the question's image,
gives the same answer as the original question.\footnote{Pruned questions may be
\emph{ill-posed} (Section~\ref{sec:question-gen}) so they are executed with
modified semantics; see supplementary material for details.} A question's
\emph{effective size} is the size of its effective question. Questions whose effective size is smaller than their actual size need not be degenerate.
The question in Figure~\ref{fig:effective-size} is not degenerate because the
entire question is needed to resolve its object references (there are two blue
cylinders and two rubber cylinders), but it has a small effective
size since it can be correctly answered without resolving those references. 

In Figure~\ref{fig:effective-size} (bottom), we show accuracy on query questions as a
function of effective question size. The error rate of all models increases
with effective question size, suggesting that models struggle with
long reasoning chains.

\begin{figure}
  \centering
  \raisebox{4mm}{
    \begin{minipage}[b]{0.187\textwidth}
      \scriptsize
      \textbf{Question:}
      \textit{There is a purple cube that is in front of the yellow metal sphere; what material is it?} \\*
      \textbf{Absolute question:}
      \textit{There is a purple cube in the front half of the image; what material is it?}
    \end{minipage}
  }
  \hspace{2mm}\includegraphics[width=0.19\textwidth]{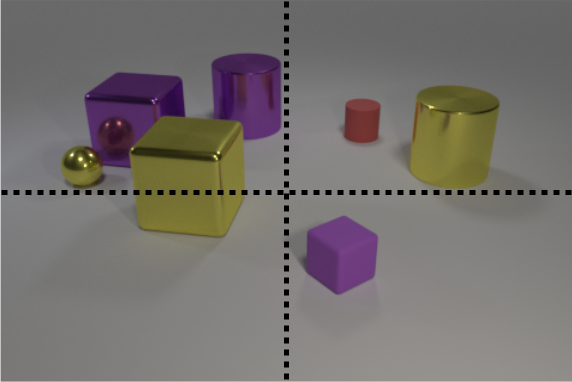}
  \includegraphics[height=0.1001\textwidth,trim={0    52px 0 0},clip]{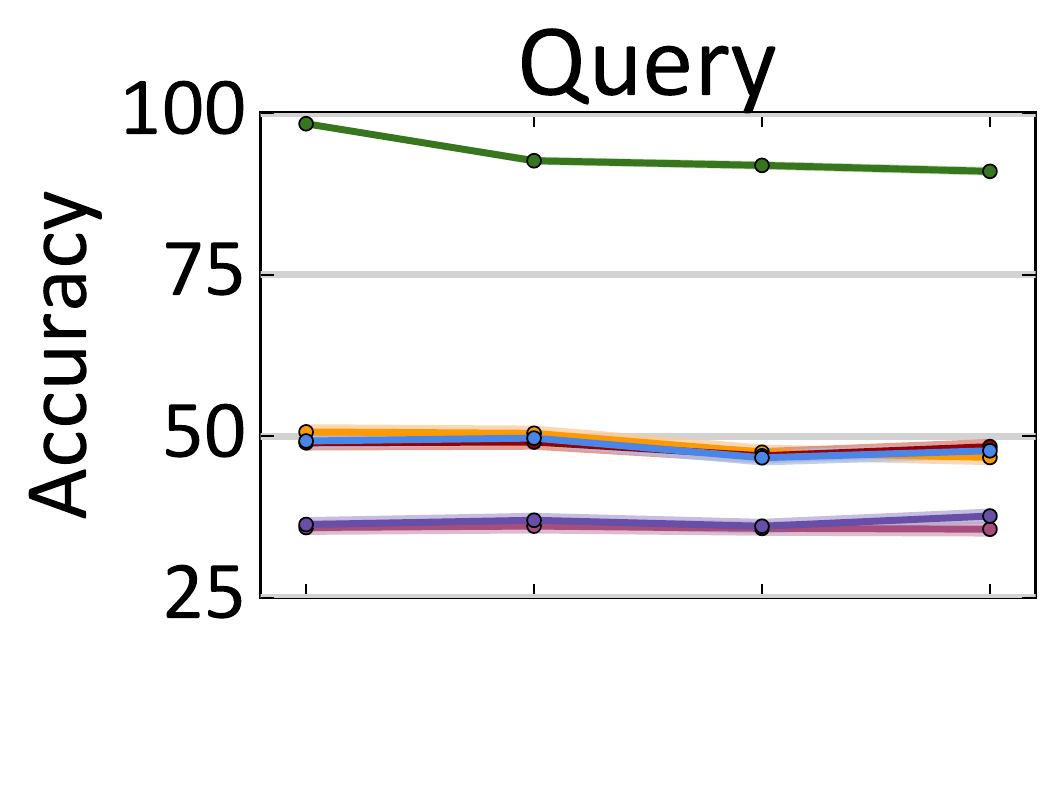}
  \includegraphics[height=0.1001\textwidth,trim={72px 52px 0 0},clip]{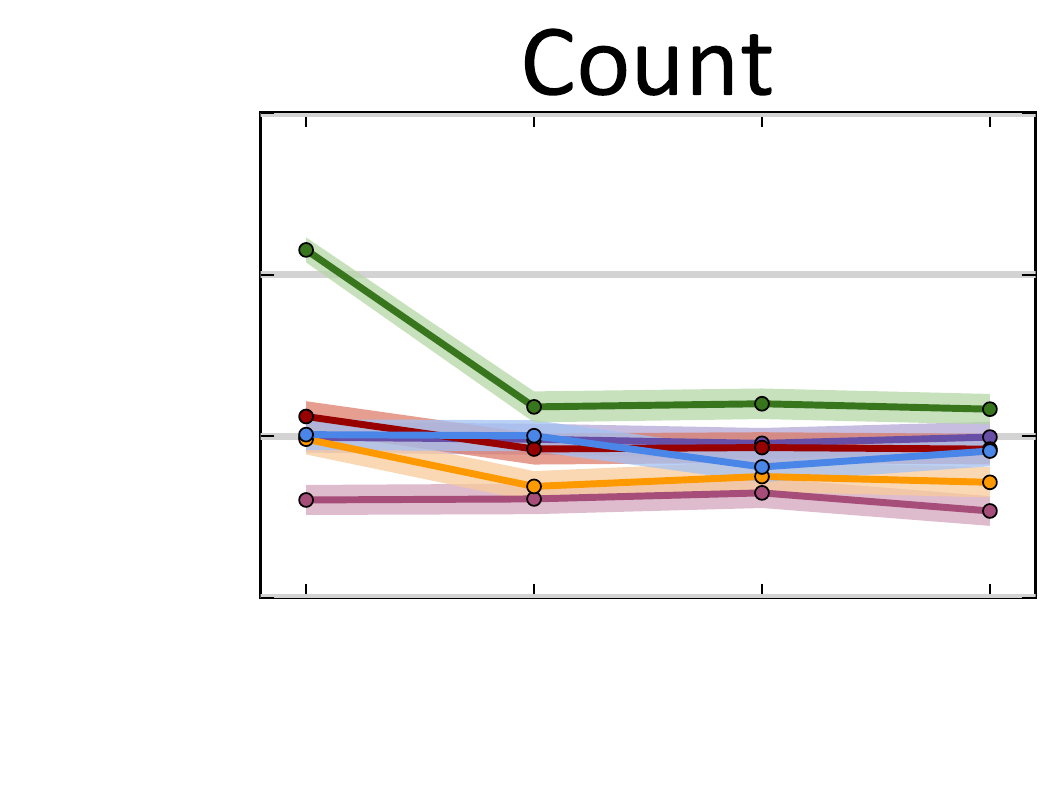}
  \includegraphics[height=0.1001\textwidth,trim={72px 52px 0 0},clip]{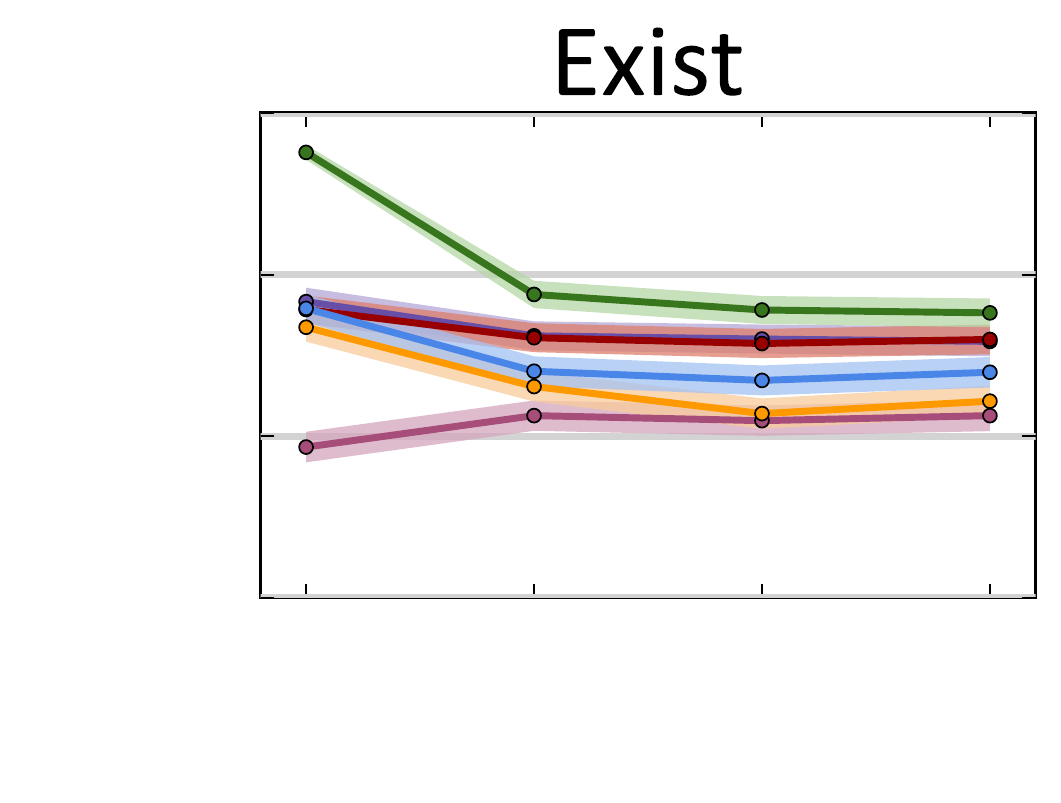}
  \includegraphics[height=0.1225\textwidth,trim={0 0 0 0px}]{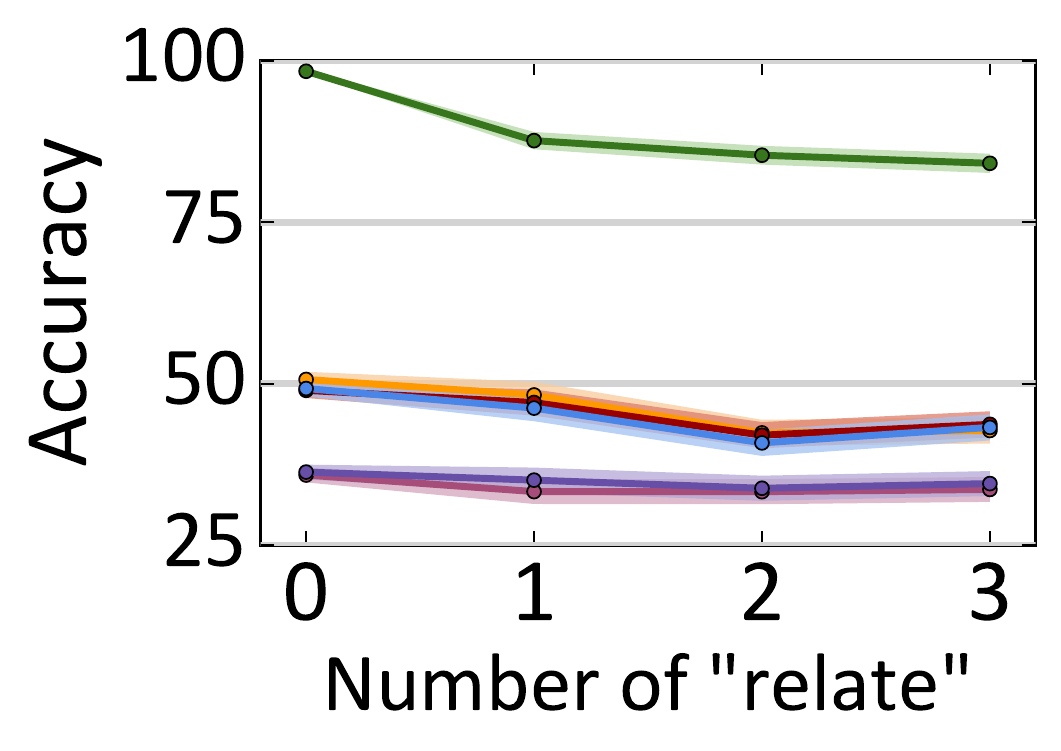}
  \includegraphics[height=0.1225\textwidth,trim={72px 0 0 0px},clip]{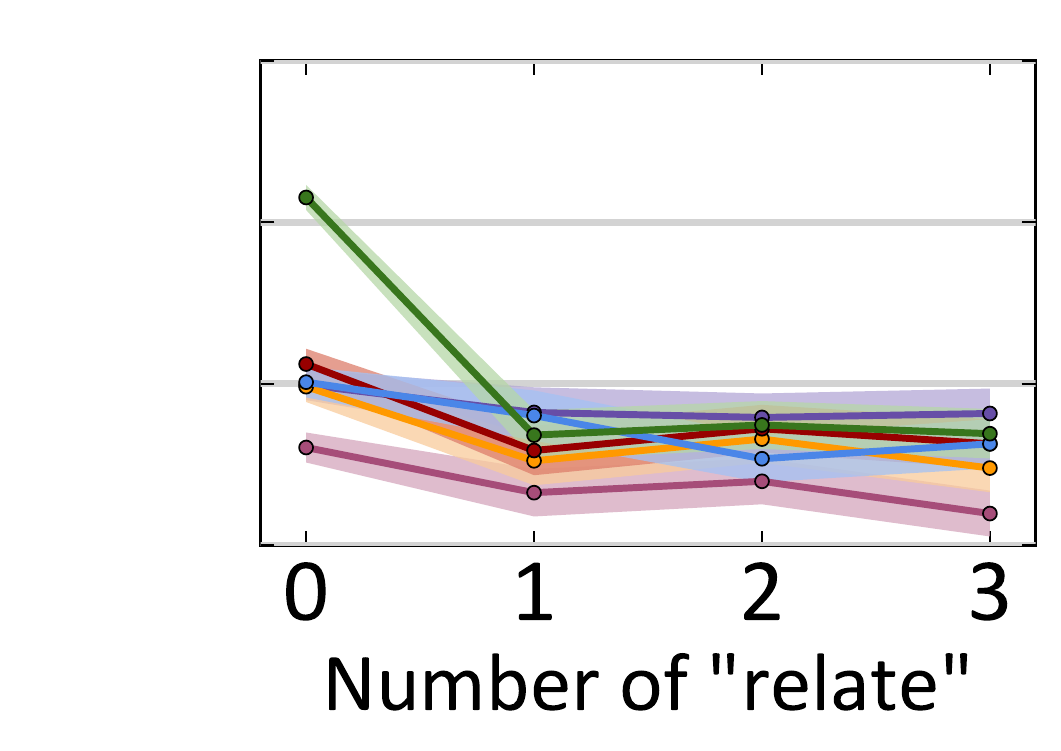}
  \includegraphics[height=0.1225\textwidth,trim={72px 0 0 0px},clip]{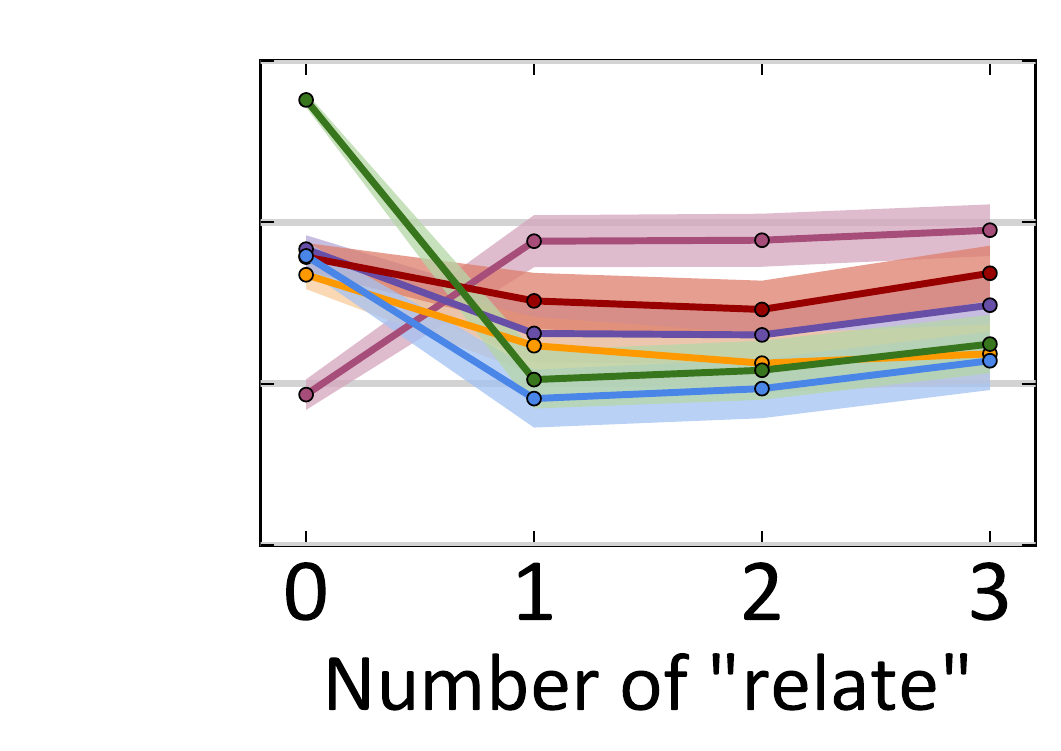} \\*
  \hspace{6mm}
  \includegraphics[width=0.4\textwidth]{figures/three_column_legend_lines.png}
  \caption{
    \textbf{Top}: Some questions can be correctly answered using \emph{absolute}
    definitions for spatial relationships; for example in this image there is only
    one purple cube in the bottom half of the image.
    \textbf{Bottom}: Accuracy of each model on \emph{chain-structured}
    questions as a function of the number of spatial relationships in the question,
    separated by question type. Top row shows all chain-structured questions;
    bottom row excludes questions that can be correctly answered using absolute
    spatial reasoning.
  }
  \vspace{-4mm}
  \label{fig:absolutespatial}
\end{figure}

\subsection{Spatial Reasoning}
\label{sec:spatial}
We expect that questions with more spatial relationships should be more challenging
since they require longer chains of reasoning. The top set of plots in
Figure~\ref{fig:absolutespatial} shows accuracy on chain-structured
questions with different numbers of relationships.\footnote{We restrict to
chain-structured questions to avoid unwanted correlations between question
topology and number of relationships.} Across all three question types, CNN+LSTM+SA
shows a significant drop in accuracy for questions with one or more spatial relationship;
other models are largely unaffected by spatial relationships.

Spatial relationships force models to reason about objects' relative positions.
However, as shown in Figure~\ref{fig:absolutespatial}, some
questions can be answered using \emph{absolute spatial reasoning}. In this question
the purple cube can be found by simply looking in the bottom half of the image;
reasoning about its position relative to the metal sphere is unnecessary.

Questions only requiring absolute spatial reasoning can be identified by modifying
the semantics of spatial relationship functions in their programs: instead of
returning sets of objects related to the input object, they ignore their input
object and return the set of objects in the half of the image corresponding to
the relationship. A question only requires absolute spatial reasoning if
executing its program with these modified semantics does not change its answer.

The bottommost plots of Figure~\ref{fig:absolutespatial} show accuracy 
on chain-structured questions with different number of relationships, \emph{excluding}
questions that can be answered with absolute spatial reasoning. On query questions,
CNN+LSTM+SA performs significantly worse when absolute spatial reasoning
is excluded; on count questions no model outperforms LSTM, and on exist questions
no model outperforms Q-type mode. These results suggest that models have not learned
the semantics of spatial relationships.

\begin{figure}
  \centering
  \includegraphics[width=0.41\textwidth]{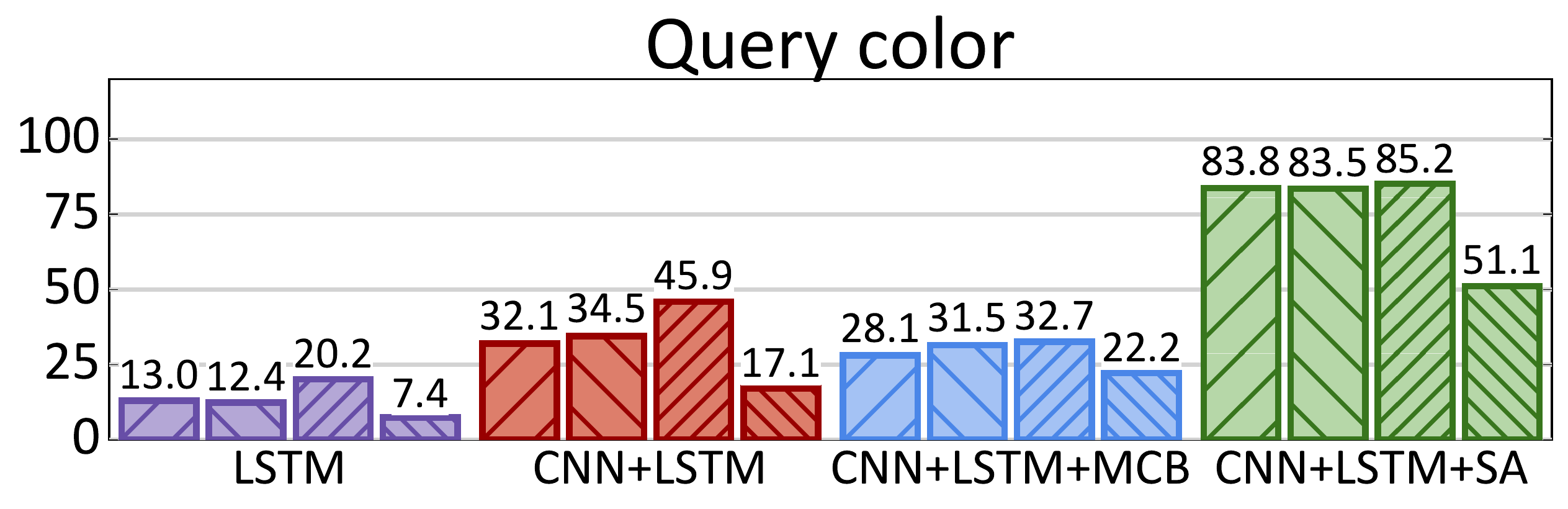}
  \includegraphics[width=0.41\textwidth]{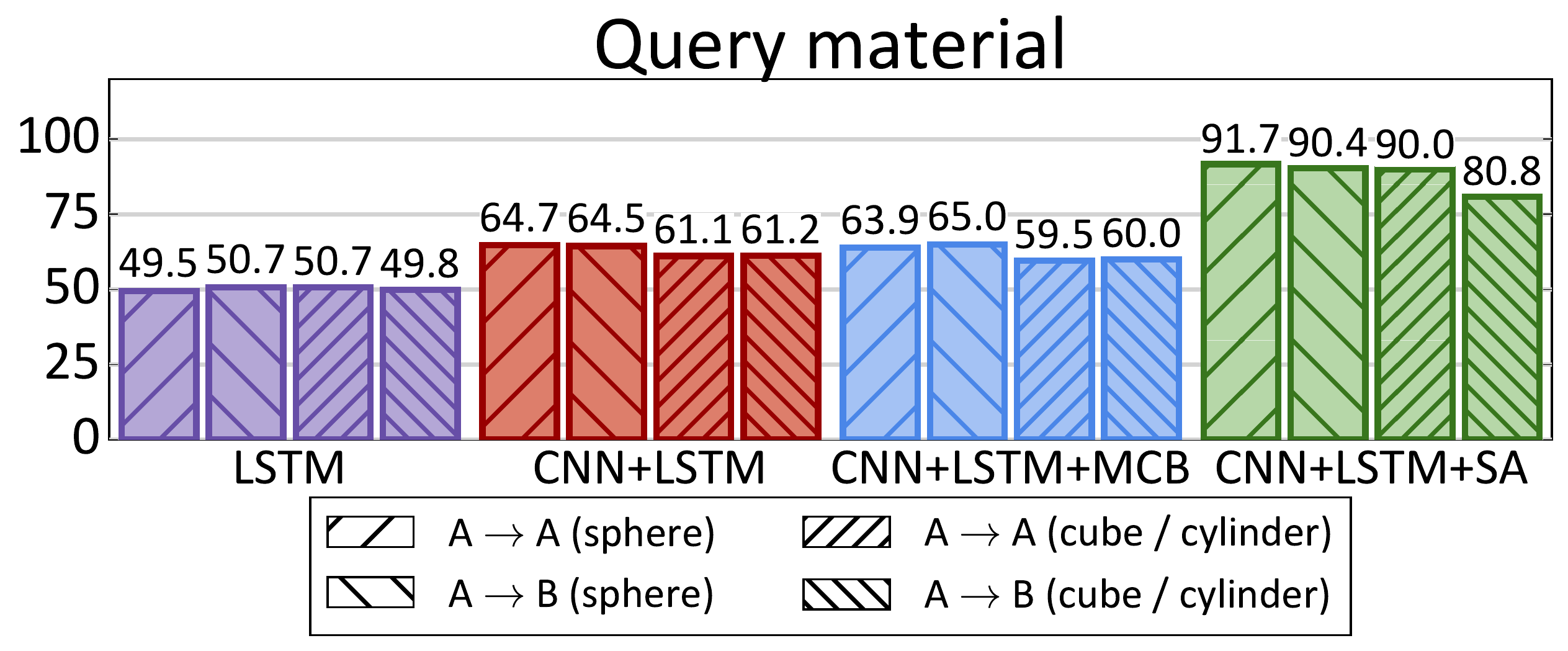}
  \caption{
    In \emph{Condition A} all cubes are gray, blue, brown, or yellow and all
    cylinders are red, green, purple, or cyan; in \emph{Condition B}  color
    palettes are swapped. We train models in Condition A and test in both
    conditions to assess their generalization performance. We show
    accuracy on ``query color'' and ``query material'' questions,
    separating questions by shape of the object being queried.
  }
  \label{fig:colorgeneralization}
  \vspace{-3mm}
\end{figure}

\subsection{Compositional Generalization}
\label{sec:disentangled}
Practical VQA systems should perform well on images and questions that contain novel
combinations of attributes not seen
during training. To do so models might need to learn disentangled representations
for attributes, for example learning separate representations for color and
shape instead of memorizing all possible color/shape combinations.

We can use CLEVR to test the ability of VQA models to perform such compositional
generalization. We synthesize two new versions of CLEVR: in \emph{Condition A}
all cubes are gray, blue, brown, or yellow and all cylinders are red, green, purple, or
cyan; in \emph{Condition B} these shapes swap color palettes.
Both conditions contain spheres of all eight colors.

We retrain models on Condition A and compare their performance when testing on Condition A
(A $\to$ A) and testing on Condition B (A $\to$ B). In Figure~\ref{fig:colorgeneralization}
we show accuracy on query-color and query-material questions, separating
questions asking about spheres (which are the same in A and B) and cubes/cylinders (which
change from A to B).

Between A$\to$A and A$\to$B, all models perform about the same when asked about
the color of spheres, but perform much worse when asked about the color of cubes
or cylinders; CNN+LSTM+SA drops from 85\% to 51\%. Models seem to learn strong
biases about the colors of objects and cannot overcome these biases when
conditions change.

When asked about the material of cubes and cylinders, CNN+LSTM+SA shows a smaller
gap between A$\to$A and A$\to$B (90\% vs 81\%); other models show no gap. Having
seen metal cubes and red metal objects during training, models can understand the
material of red metal cubes.

\section{Discussion and Future Work}
This paper has introduced CLEVR, a dataset designed to aid in diagnostic
evaluation of visual question answering (VQA) systems by minimizing
dataset bias and providing rich ground-truth representations for both
images and questions. Our experiments demonstrate that CLEVR facilitates
in-depth analysis not possible with other VQA datasets: our question
representations allow us to slice the dataset along different axes
(question type, relationship type, question topology, \etc), and comparing
performance along these different axes allows us to better
understand the reasoning capabilities of VQA systems.
Our analysis has revealed several key shortcomings of current VQA systems:

\vspace{-4pt}
\begin{packed_item}
  \item\textbf{Short-term memory}: All systems we tested performed poorly
  in situations requiring short-term memory, including attribute comparison
  and integer equality questions (Section~\ref{sec:question-type}), same-attribute 
  relationships (Section~\ref{sec:relationship-type}), and tree-structured
  questions (Section~\ref{sec:topology}). Attribute comparison questions are
  of particular interest, since models can successfully \emph{identity} attributes
  of objects but struggle to \emph{compare} attributes.
  \item\textbf{Long reasoning chains}: Systems struggle to answer questions
  requiring long chains of nontrivial reasoning, including questions with
  large effective sizes (Section~\ref{sec:effective-size}) and count and
  existence questions with many spatial relationships (Section~\ref{sec:spatial}).
  \item\textbf{Spatial Relationships}: Models fail
  to learn the true semantics of spatial relationships, instead relying on
  absolute image position (Section~\ref{sec:spatial}).
  \item\textbf{Disentangled Representations}: By training and testing models
  on different data distributions (Section~\ref{sec:disentangled}) we argue
  that models do not learn representations
  that properly disentangle object attributes; they seem to learn strong biases
  from the training data and cannot overcome these biases when conditions change.
\end{packed_item}
\vspace{-4pt}

Our study also shows cases where current VQA systems are successful.
In particular, spatial attention~\cite{yang16} allows models
to focus on objects and identify their attributes even on questions requiring
multiple steps of reasoning.

These observations present clear avenues for future work on VQA. We plan to use CLEVR to study models with explicit short-term memory, facilitating comparisons between values
\cite{graves16,joulin15b,weston15,xiong2016dynamic}; explore approaches that encourage learning disentangled representations \cite{bengio14}; and investigate methods that compile custom network architectures for different patterns of reasoning~\cite{andreas16b,andreas16}. We hope that diagnostic datasets like CLEVR will help guide future research in VQA and enable rapid progress on this important task.

\clearpage
\pagebreak
\appendix
\begin{figure}[ht!]
  \centering
  \Large\textbf{Supplementary Material}
\end{figure}

\section{Basic Functions}
As described in Section 3 and shown in Figure 2, each question in CLEVR
is associated with a functional program built from a set of basic functions.
In this section we detail the semantics of these basic functional building
blocks.

\paragraph{Data Types.}
Our basic functional building blocks operate on values of the following types:
\begin{packed_item}
  \item \texttt{Object}: A single object in the scene.
  \item \texttt{ObjectSet}: A set of zero or more objects in the scene.
  \item \texttt{Integer}: An integer between 0 and 10 (inclusive).
  \item \texttt{Boolean}: Either \texttt{yes} or \texttt{no}.
  \item Value types:
  \begin{packed_item}
    \item \texttt{Size}: One of \texttt{large} or \texttt{small}.
    \item \texttt{Color}: One of \texttt{gray}, \texttt{red}, \texttt{blue}, \texttt{green}, \texttt{brown}, \texttt{purple}, \texttt{cyan}, or \texttt{yellow}.
    \item \texttt{Shape}: One of \texttt{cube}, \texttt{sphere}, or \texttt{cylinder}.
    \item \texttt{Material}: One of \texttt{rubber} or \texttt{metal}.
  \end{packed_item}
  \item \texttt{Relation}: One of \texttt{left}, \texttt{right},
        \texttt{in front}, or \texttt{behind}.
\end{packed_item}

\paragraph{Basic Functions.}
The functional program representations of questions are built from the following
set of basic building blocks. Each of these functions takes the image's scene graph
as an additional implicit input.
\begin{packed_item}
  \item \textbf{scene}
    {\footnotesize ($\emptyset\to\texttt{ObjectSet}$)} \\*
    Returns the set of all objects in the scene.
  \item \textbf{unique}
    {\footnotesize ($\texttt{ObjectSet}\to\texttt{Object}$)} \\*
    If the input is a singleton set, then return it as a standalone
    \texttt{Object}; otherwise raise an exception and flag the question
    as \emph{ill-posed} (See Section 3).
  \item \textbf{relate}
    {\footnotesize ($\texttt{Object}\times\texttt{Relation}\to\texttt{ObjectSet}$)} \\*
    Return all objects in the scene that have the specified spatial relation to the input
    object. For example if the input object is a red cube and the input relation is
    \texttt{left}, then return the set of all objects in the scene that are left of the
    red cube.
  \item \textbf{count}
    {\footnotesize ($\texttt{ObjectSet}\to\texttt{Integer}$)} \\*
    Returns the size of the input set.
  \item \textbf{exist}
    {\footnotesize ($\texttt{ObjectSet}\to\texttt{Boolean}$)} \\*
    Returns \texttt{yes} if the input set is nonempty and \texttt{no} if it is empty.
  \item \textbf{Filtering functions}: These functions filter the input objects by some attribute,
    returning the subset of input objects that match the input attribute. For example
    calling \texttt{filter\_size} with the first input \texttt{small} will return the
    set of all small objects in the second input.
  \begin{packed_item}
    \item \textbf{filter\_size}
      {\footnotesize ($\texttt{ObjectSet}\times\texttt{Size}\to\texttt{ObjectSet}$)}
    \item \textbf{filter\_color}
      {\footnotesize ($\texttt{ObjectSet}\times\texttt{Color}\to\texttt{ObjectSet}$)}
    \item \textbf{filter\_material}
      {\scriptsize ($\texttt{ObjectSet}\times\texttt{Material}\to\texttt{ObjectSet}$)}
    \item \textbf{filter\_shape}
      {\footnotesize ($\texttt{ObjectSet}\times\texttt{Shape}\to\texttt{ObjectSet}$)}
  \end{packed_item}
  \item \textbf{Query functions}: These functions return the specified attribute of the input
  object; for example calling \texttt{query\_color} on a red object returns \texttt{red}.
  \begin{packed_item}
    \item \textbf{query\_size}
      {\footnotesize ($\texttt{Object}\to\texttt{Size}$)}
    \item \textbf{query\_color}
      {\footnotesize ($\texttt{Object}\to\texttt{Color}$)}
    \item \textbf{query\_material}
      {\footnotesize ($\texttt{Object}\to\texttt{Material}$)}
    \item \textbf{query\_shape}
      {\footnotesize ($\texttt{Object}\to\texttt{Shape}$)}
  \end{packed_item}
  \item \textbf{Logical operators}:
  \begin{packed_item}
    \item \textbf{AND}
      {\footnotesize ($\texttt{ObjectSet}\times\texttt{ObjectSet}\to\texttt{ObjectSet}$)} \\*
      Returns the intersection of the two input sets.
    \item \textbf{OR}
      {\footnotesize ($\texttt{ObjectSet}\times\texttt{ObjectSet}\to\texttt{ObjectSet}$)} \\*
      Returns the union of the two input sets.
  \end{packed_item}
  \item \textbf{Same-attribute relations}: These functions return the set of
    objects that have the same attribute value as the input object, not
    including the input object. For example calling \texttt{same\_shape} on
    a cube returns the set of all cubes in the scene, excluding the query cube.
  \begin{packed_item}
    \item \textbf{same\_size}
      {\footnotesize ($\texttt{Object}\to\texttt{ObjectSet}$)}
    \item \textbf{same\_color}
      {\footnotesize ($\texttt{Object}\to\texttt{ObjectSet}$)}
    \item \textbf{same\_material}
      {\footnotesize ($\texttt{Object}\to\texttt{ObjectSet}$)}
    \item \textbf{same\_shape}
      {\footnotesize ($\texttt{Object}\to\texttt{ObjectSet}$)}
  \end{packed_item}
  \item \textbf{Integer comparison}: Checks whether the two integer inputs are
    equal, or whether the first is less than or greater than the second, returning
    either \texttt{yes} or \texttt{no}.
  \begin{packed_item}
    \item \textbf{equal\_integer}
      {\footnotesize ($\texttt{Integer}\times\texttt{Integer}\to\texttt{Boolean})$}
    \item \textbf{less\_than}
      {\footnotesize ($\texttt{Integer}\times\texttt{Integer}\to\texttt{Boolean})$}
    \item \textbf{greater\_than}
      {\footnotesize ($\texttt{Integer}\times\texttt{Integer}\to\texttt{Boolean})$}
  \end{packed_item}
  \item \textbf{Attribute comparison}: These functions return \texttt{yes} if their
    inputs are equal and \texttt{no} if they are not equal.
  \begin{packed_item}
    \item \textbf{equal\_size}
    {\footnotesize ($\texttt{Size}\times\texttt{Size}\to\texttt{Boolean})$}
    \item \textbf{equal\_material}
    {\scriptsize ($\texttt{Material}\times\texttt{Material}\to\texttt{Boolean})$}
    \item \textbf{equal\_color}
    {\footnotesize ($\texttt{Color}\times\texttt{Color}\to\texttt{Boolean})$}
    \item \textbf{equal\_shape}
    {\footnotesize ($\texttt{Shape}\times\texttt{Shape}\to\texttt{Boolean})$}
  \end{packed_item}
\end{packed_item}

\section{Effective Question Size}
In Section 4.5 we note that some questions can be correctly answered without
correctly resolving all intermediate object references, and define a question's
\emph{effective question} to quantitatively measure this effect.

For any question we can compute its \emph{effective question} by pruning functions
from the question's program; the effective question is the smallest such pruned
program that, when executed on the scene graph for the question's image, gives
the same answer as the original question.

\begin{figure}
\includegraphics[width=0.45\textwidth]{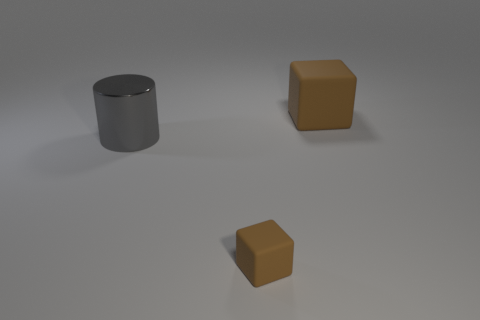}
\caption{For the above image and the question ``What color is the cube behind the cylinder?", the effective question is ``What color is the cube?" (see text for details).}
\label{fig:supp:effectivesize}
\end{figure}
Some pruned questions may be \emph{ill-posed}, meaning that some object
references do not refer to a unique object. For example, consider the question
\textit{``What color is the cube behind the cylinder?''}; its
associated program is
\setlength{\abovedisplayskip}{3pt}
\setlength{\belowdisplayskip}{3pt}
\begin{multline*}
  \emph{query\_color(unique(filter\_shape(cube,} \\
    \emph{relate(behind, unique(filter\_shape(cylinder, scene()))))))}
\end{multline*}

Imagine executing this program on the scene shown in Figure~\ref{fig:supp:effectivesize}. 
The innermost \emph{filter\_shape} gives a set containing the cylinder, the \emph{relate}
returns a set containing just the large cube in the back, the outer
\emph{filter\_shape} does nothing, and the \emph{query\_color} returns
\texttt{brown}.

This question is not \emph{ill-posed} (Section 3)
because the reference to \emph{``the cube''} cannot be resolved without
the rest of the question; however this question's effective size is less
than its actual size because the question can be correctly answered
without resolving this object reference correctly.

To compute the effective question, we attempt to prune functions from this
program. Starting from the innermost function and working out, whenever we
find a function whose input type is \texttt{Object} or \texttt{ObjectSet},
we construct a pruned question by replacing that function's input with a
\textit{scene} function and executing it. The smallest such pruned program
that gives the same answer as the original program is the effective question.

Pruned questions may be ill-posed, so we execute them with modified semantics.
The output type of the \emph{unique} function is changed from \texttt{Object}
to \texttt{ObjectSet}, and it simply returns its input set. All functions
taking an \texttt{Object} input are modified to take an \texttt{ObjectSet}
input instead by mapping the original function over its input set and flattening
the resulting set; thus the \emph{relate} functions return the set of objects
in the scene that have the specified relationship with any of the input
objects, and the \emph{query} functions return sets of values rather than
single values.

Therefore for this example question we consider the following sequence
of pruned programs.
First we prune the inner \emph{filter\_shape} function:
\begin{multline*}
  \emph{query\_color(unique(filter\_shape(cube,} \\
    \emph{relate(behind, unique(scene())))))}
\end{multline*}
The \emph{relate} function now returns the set of objects which are
behind some object, so it returns the large cube and the cylinder
 (since it is behind the small cube). The
\emph{filter\_shape} function removes the cylinder, and the \emph{query\_color}
returns a singleton set containing \texttt{brown}.

Next we prune the inner \emph{unique} function:
\begin{multline*}
  \emph{query\_color(unique(filter\_shape(cube,} \\
    \emph{relate(behind, scene()))))}
\end{multline*}
Since \emph{unique} computes the identity for pruned questions, execution
is the same as above.

Next we prune the \emph{relate} function:
\begin{multline*}
  \emph{query\_color(unique(filter\_shape(cube, scene())))}
\end{multline*}
Now the \emph{filter\_shape} returns the set of both cubes,
but since both are brown the \emph{query\_color} still returns a
singleton set containing \texttt{brown}.

Next we prune the \emph{filter\_shape} function:
\begin{multline*}
  \emph{query\_color(unique(scene()))}
\end{multline*}
Now the \emph{query\_color} receives the set of all three input
objects, so it returns a set containing \texttt{brown} and
\texttt{gray}, which is different from the original question.

The effective question is therefore:
\begin{multline*}
  \emph{query\_color(unique(filter\_shape(cube, scene())))}
\end{multline*}

\begin{figure}
  \centering
  \includegraphics[height=0.137\textwidth,trim={0 0 0 0},clip]{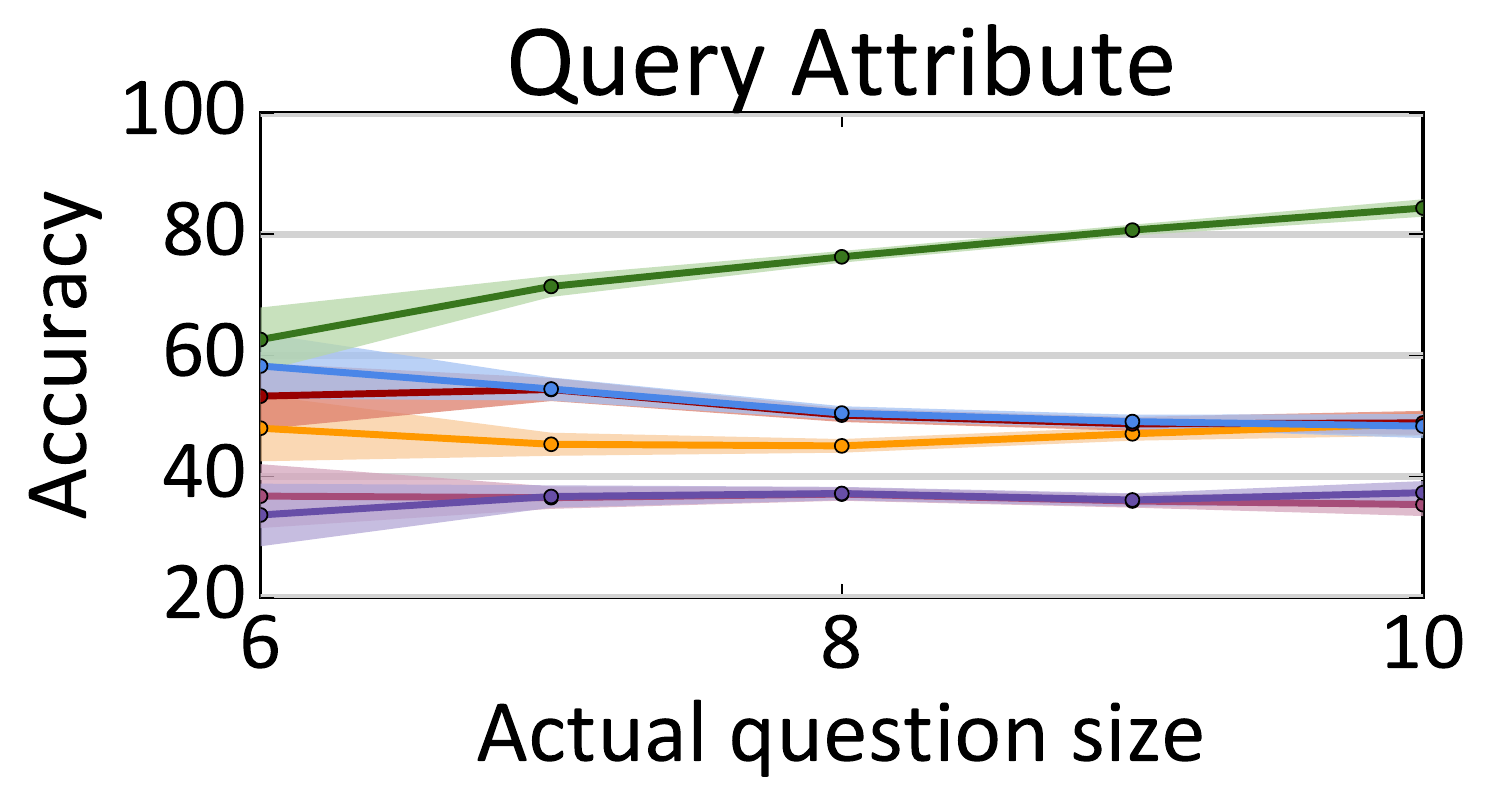}
  \includegraphics[height=0.137\textwidth,trim={70px 0 0 0},clip]{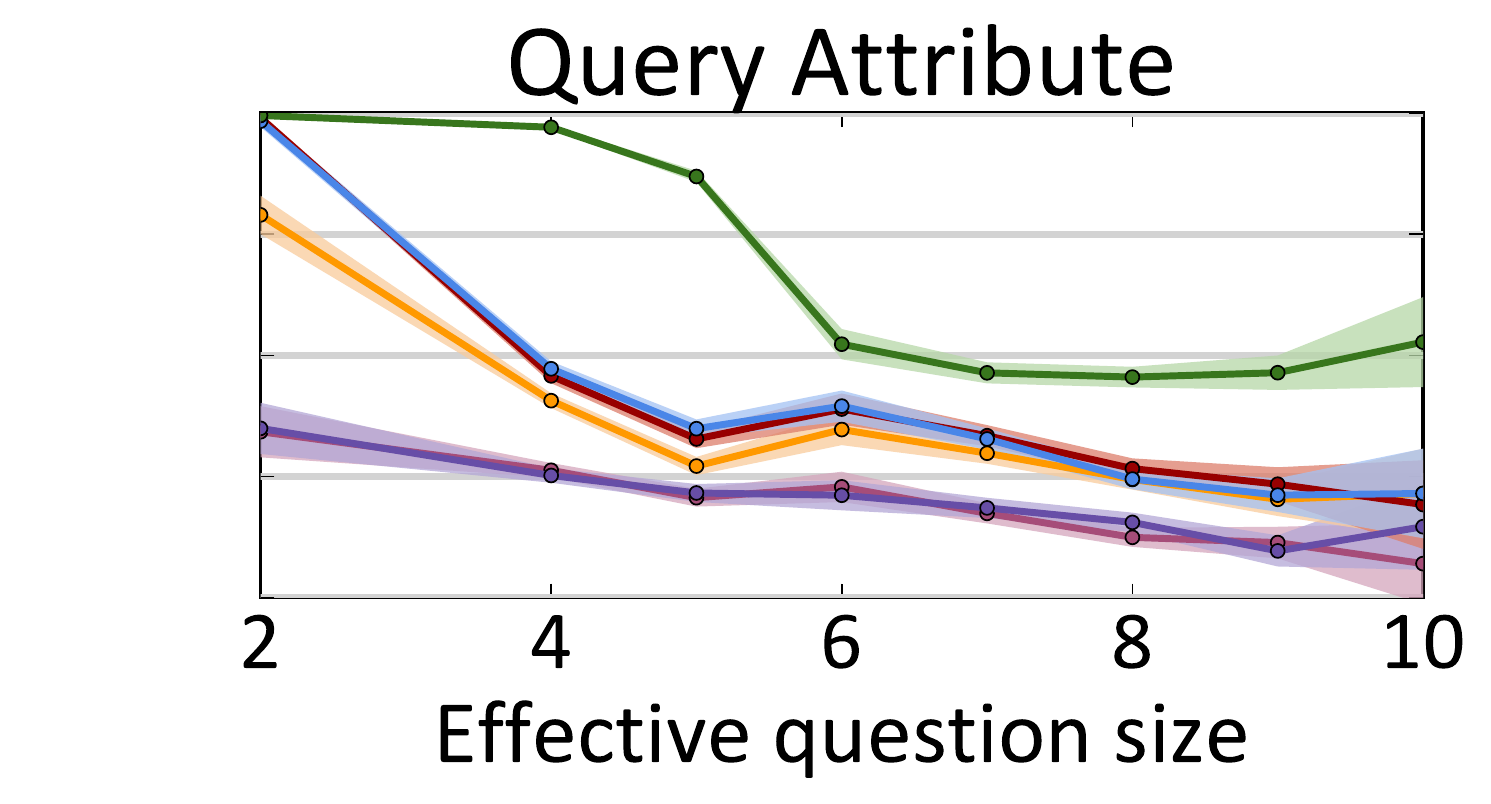} \\*
  \hspace{15px}\includegraphics[width=0.4\textwidth]{figures/three_column_legend_lines.png}
  \caption{
    Accuracy on query questions \vs actual and effective question size, restricting to questions
    with a \emph{same-attribute} relationship. Figure 7 shows the same plots for questions
    without a \emph{same-attribute} relationship. For both groups of questions we see that
    accuracy decreases as effective question size increases.
  }
  \label{fig:effective-size-with-same}
\end{figure}

\subsection{Accuracy vs Question Size}
Figure 7 of the main paper shows model accuracy on \emph{query-attribute} questions as
a function of actual and effective question size, excluding questions with
\emph{same-attribute} relationships. Questions with same-attribute relationships have
a maximum question size of 10 but questions without same-attribute relationships have
a maximum size of 20; combining these questions thus leads to unwanted correlations
between question size and difficulty.

In Figure~\ref{fig:effective-size-with-same} we show model accuracy \vs actual and
effective question size for questions with same-attribute relationships. Similar to
Figure 7, we see that model accuracy either remains constant or increases as actual
question size increases, but all models show a clear decrease in accuracy as effective
question size increases.

\section{Dynamic Module Networks}
Module networks~\cite{andreas16b,andreas16} are a novel approach to visual question
answering where a set of differentiable \emph{modules} are used to assemble a custom
network architecture to answer each question. Each module is responsible for
performing a specific function such as \emph{finding} a particular type of object,
\emph{describing} the current object of attention, or performing a
\emph{logical and} operation to merge attention masks. This approach seems
like a natural fit for the rich, compositional questions in CLEVR; unfortunately
we found that parsing heuristics tuned for the VQA dataset did not generalize
to the longer, more complex questions in CLEVR.

Dynamic module networks~\cite{andreas16b} generate network architectures by
performing a dependency parse of the question, using a set of heuristics
to compute a set of \emph{layout fragments}, combining these fragments to
create \emph{candidate layouts}, and ranking the candidate layouts using
an MLP.

For some questions, the heuristics are unable to produce any layout fragments;
in this case, the system uses a simple default network architecture as a fallback
for answering that question. On a random sample of 10,000 questions from the VQA
dataset~\cite{antol15}, we found that dynamic module networks resorted to default
architecture for 7.8\% of questions; on a random sample of 10,000 questions from
CLEVR, the default network architecture was used for 28.9\% of questions. This
suggests that the same parsing heuristics used for VQA do not apply to the
questions in CLEVR; therefore the method of~\cite{andreas16b} did not work
out-of-the box on CLEVR.

\section{Example images and questions}
The remaining pages show randomly selected images and questions from CLEVR.
Each question is annotated with its answer, question type, and size.
Recall from Section 3 that a question's \emph{type} is the outermost function
in the question's functional program, and a question's \emph{size} is the
number of functions in its program.

\begin{figure*}  \begin{minipage}{0.32\textwidth}
    \includegraphics[width=\textwidth]{figures/example_images/047090.png}
    \begin{minipage}[t][4cm][t]{0.46\textwidth}
      \footnotesize
      \textbf{Q:} There is a rubber cube in front of the big cylinder in front of the big brown matte thing; what is its size? \\*
      \textbf{A:} small \\*
      \textbf{Q-type:} query\_size \\*
      \textbf{Size:} 14 \\*[6pt]
    \end{minipage}\hspace{1mm}
    \begin{minipage}[t][4cm][t]{0.46\textwidth}
      \footnotesize
      \textbf{Q:} What color is the object that is on the left side of the small rubber thing? \\*
      \textbf{A:} gray \\*
      \textbf{Q-type:} query\_color \\*
      \textbf{Size:} 7 \\*[6pt]
    \end{minipage}\\*
    \begin{minipage}[t][4cm][t]{0.46\textwidth}
      \footnotesize
      \textbf{Q:} There is another cube that is made of the same material as the small brown block; what is its size? \\*
      \textbf{A:} large \\*
      \textbf{Q-type:} query\_size \\*
      \textbf{Size:} 9 \\*[6pt]
    \end{minipage}\hspace{1mm}
    \begin{minipage}[t][4cm][t]{0.46\textwidth}
      \footnotesize
      \textbf{Q:} What number of other matte objects are the same shape as the small rubber object? \\*
      \textbf{A:} 1 \\*
      \textbf{Q-type:} count \\*
      \textbf{Size:} 7 \\*[6pt]
    \end{minipage}
  \end{minipage}
  \hspace{1mm}
  \begin{minipage}{0.32\textwidth}
    \includegraphics[width=\textwidth]{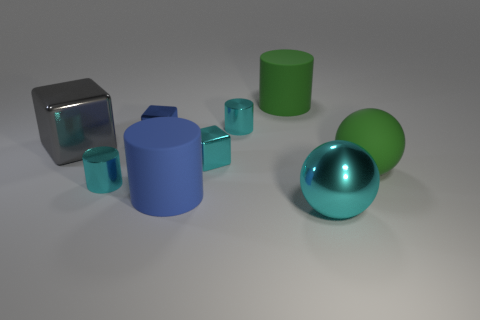}
    \begin{minipage}[t][4cm][t]{0.46\textwidth}
      \footnotesize
      \textbf{Q:} Are there fewer metallic objects that are on the left side of the large cube than cylinders to the left of the cyan shiny block? \\*
      \textbf{A:} yes \\*
      \textbf{Q-type:} less\_than \\*
      \textbf{Size:} 16 \\*[6pt]
    \end{minipage}\hspace{1mm}
    \begin{minipage}[t][4cm][t]{0.46\textwidth}
      \footnotesize
      \textbf{Q:} There is a green rubber thing that is left of the rubber thing that is right of the rubber cylinder behind the gray shiny block; what is its size? \\*
      \textbf{A:} large \\*
      \textbf{Q-type:} query\_size \\*
      \textbf{Size:} 17 \\*[6pt]
    \end{minipage}\\*
    \begin{minipage}[t][4cm][t]{0.46\textwidth}
      \footnotesize
      \textbf{Q:} What is the size of the matte thing that is on the left side of the large cyan object and in front of the small blue thing? \\*
      \textbf{A:} large \\*
      \textbf{Q-type:} query\_size \\*
      \textbf{Size:} 14 \\*[6pt]
    \end{minipage}\hspace{1mm}
    \begin{minipage}[t][4cm][t]{0.46\textwidth}
      \footnotesize
      \textbf{Q:} The green matte thing that is behind the large green thing that is to the right of the big cyan metal object is what shape? \\*
      \textbf{A:} cylinder \\*
      \textbf{Q-type:} query\_shape \\*
      \textbf{Size:} 14 \\*[6pt]
    \end{minipage}
  \end{minipage}
  \hspace{1mm}
  \begin{minipage}{0.32\textwidth}
    \includegraphics[width=\textwidth]{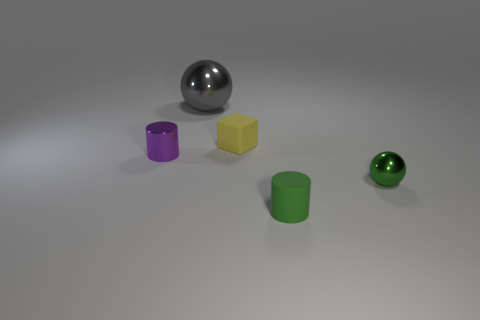}
    \begin{minipage}[t][4cm][t]{0.46\textwidth}
      \footnotesize
      \textbf{Q:} Is the number of tiny metal objects left of the purple metallic cylinder the same as the number of small metal objects to the left of the small metal sphere? \\*
      \textbf{A:} no \\*
      \textbf{Q-type:} equal\_integer \\*
      \textbf{Size:} 19 \\*[6pt]
    \end{minipage}\hspace{1mm}
    \begin{minipage}[t][4cm][t]{0.46\textwidth}
      \footnotesize
      \textbf{Q:} Do the large shiny object and the thing to the left of the big gray object have the same shape? \\*
      \textbf{A:} no \\*
      \textbf{Q-type:} equal\_shape \\*
      \textbf{Size:} 13 \\*[6pt]
    \end{minipage}\\*
    \begin{minipage}[t][4cm][t]{0.46\textwidth}
      \footnotesize
      \textbf{Q:} There is a small thing that is the same color as the tiny shiny ball; what is it made of? \\*
      \textbf{A:} rubber \\*
      \textbf{Q-type:} query\_material \\*
      \textbf{Size:} 9 \\*[6pt]
    \end{minipage}\hspace{1mm}
    \begin{minipage}[t][4cm][t]{0.46\textwidth}
      \footnotesize
      \textbf{Q:} Is there anything else that is the same shape as the tiny yellow matte thing? \\*
      \textbf{A:} no \\*
      \textbf{Q-type:} exist \\*
      \textbf{Size:} 7 \\*[6pt]
    \end{minipage}
  \end{minipage}
  \begin{minipage}{0.32\textwidth}
    \includegraphics[width=\textwidth]{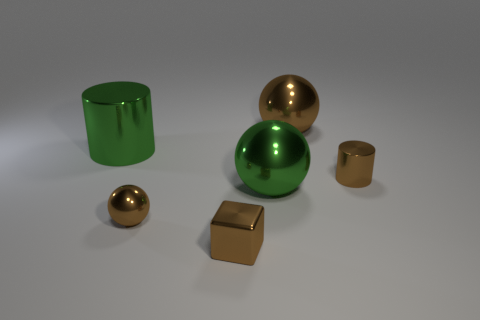}
    \begin{minipage}[t][4cm][t]{0.46\textwidth}
      \footnotesize
      \textbf{Q:} What color is the small shiny cube? \\*
      \textbf{A:} brown \\*
      \textbf{Q-type:} query\_color \\*
      \textbf{Size:} 6 \\*[6pt]
    \end{minipage}\hspace{1mm}
    \begin{minipage}[t][4cm][t]{0.46\textwidth}
      \footnotesize
      \textbf{Q:} There is a tiny shiny sphere left of the cylinder in front of the large cylinder; what color is it? \\*
      \textbf{A:} brown \\*
      \textbf{Q-type:} query\_color \\*
      \textbf{Size:} 13 \\*[6pt]
    \end{minipage}\\*
    \begin{minipage}[t][4cm][t]{0.46\textwidth}
      \footnotesize
      \textbf{Q:} Is there a tiny thing that has the same material as the brown cylinder? \\*
      \textbf{A:} yes \\*
      \textbf{Q-type:} exist \\*
      \textbf{Size:} 7 \\*[6pt]
    \end{minipage}\hspace{1mm}
    \begin{minipage}[t][4cm][t]{0.46\textwidth}
      \footnotesize
      \textbf{Q:} What is the material of the tiny object to the right of the brown shiny ball behind the tiny shiny cylinder? \\*
      \textbf{A:} metal \\*
      \textbf{Q-type:} query\_material \\*
      \textbf{Size:} 14 \\*[6pt]
    \end{minipage}
  \end{minipage}
  \hspace{1mm}
  \begin{minipage}{0.32\textwidth}
    \includegraphics[width=\textwidth]{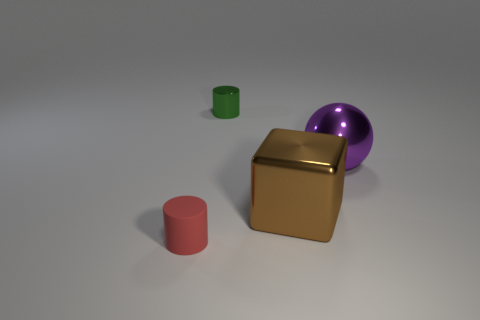}
    \begin{minipage}[t][4cm][t]{0.46\textwidth}
      \footnotesize
      \textbf{Q:} What number of cylinders are purple metal objects or purple matte things? \\*
      \textbf{A:} 0 \\*
      \textbf{Q-type:} count \\*
      \textbf{Size:} 9 \\*[6pt]
    \end{minipage}\hspace{1mm}
    \begin{minipage}[t][4cm][t]{0.46\textwidth}
      \footnotesize
      \textbf{Q:} Does the large purple shiny object have the same shape as the tiny object that is behind the matte thing? \\*
      \textbf{A:} no \\*
      \textbf{Q-type:} equal\_shape \\*
      \textbf{Size:} 14 \\*[6pt]
    \end{minipage}\\*
    \begin{minipage}[t][4cm][t]{0.46\textwidth}
      \footnotesize
      \textbf{Q:} There is a object that is both on the left side of the brown metal block and in front of the large purple shiny ball; how big is it? \\*
      \textbf{A:} small \\*
      \textbf{Q-type:} query\_size \\*
      \textbf{Size:} 16 \\*[6pt]
    \end{minipage}\hspace{1mm}
    \begin{minipage}[t][4cm][t]{0.46\textwidth}
      \footnotesize
      \textbf{Q:} There is a big purple shiny object; what shape is it? \\*
      \textbf{A:} sphere \\*
      \textbf{Q-type:} query\_shape \\*
      \textbf{Size:} 6 \\*[6pt]
    \end{minipage}
  \end{minipage}
  \hspace{1mm}
  \begin{minipage}{0.32\textwidth}
    \includegraphics[width=\textwidth]{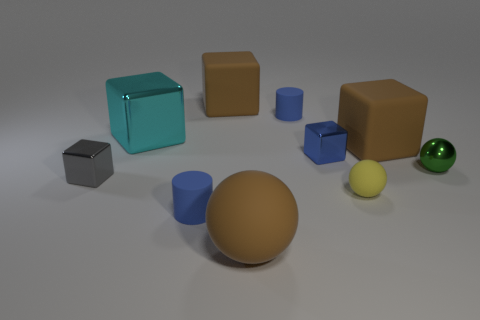}
    \begin{minipage}[t][4cm][t]{0.46\textwidth}
      \footnotesize
      \textbf{Q:} There is a large brown block in front of the tiny rubber cylinder that is behind the cyan block; are there any big cyan metallic cubes that are to the left of it? \\*
      \textbf{A:} yes \\*
      \textbf{Q-type:} exist \\*
      \textbf{Size:} 20 \\*[6pt]
    \end{minipage}\hspace{1mm}
    \begin{minipage}[t][4cm][t]{0.46\textwidth}
      \footnotesize
      \textbf{Q:} There is a big shiny object; are there any blue shiny cubes behind it? \\*
      \textbf{A:} no \\*
      \textbf{Q-type:} exist \\*
      \textbf{Size:} 9 \\*[6pt]
    \end{minipage}\\*
    \begin{minipage}[t][4cm][t]{0.46\textwidth}
      \footnotesize
      \textbf{Q:} What number of cylinders have the same color as the metal ball? \\*
      \textbf{A:} 0 \\*
      \textbf{Q-type:} count \\*
      \textbf{Size:} 7 \\*[6pt]
    \end{minipage}\hspace{1mm}
    \begin{minipage}[t][4cm][t]{0.46\textwidth}
      \footnotesize
      \textbf{Q:} The cyan block that is the same material as the tiny gray thing is what size? \\*
      \textbf{A:} large \\*
      \textbf{Q-type:} query\_size \\*
      \textbf{Size:} 9 \\*[6pt]
    \end{minipage}
  \end{minipage}
\end{figure*}
\begin{figure*}
  \begin{minipage}{0.32\textwidth}
    \includegraphics[width=\textwidth]{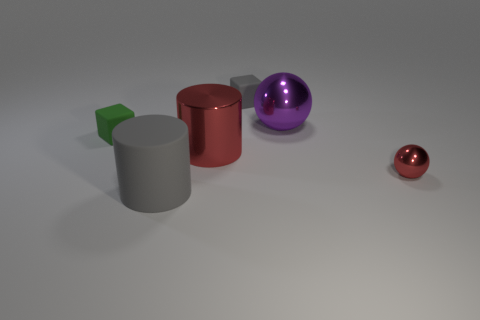}
    \begin{minipage}[t][4cm][t]{0.46\textwidth}
      \footnotesize
      \textbf{Q:} How big is the gray rubber object that is behind the big shiny thing behind the big metallic thing that is on the left side of the purple ball? \\*
      \textbf{A:} small \\*
      \textbf{Q-type:} query\_size \\*
      \textbf{Size:} 17 \\*[6pt]
    \end{minipage}\hspace{1mm}
    \begin{minipage}[t][4cm][t]{0.46\textwidth}
      \footnotesize
      \textbf{Q:} There is a purple ball that is the same size as the red cylinder; what material is it? \\*
      \textbf{A:} metal \\*
      \textbf{Q-type:} query\_material \\*
      \textbf{Size:} 9 \\*[6pt]
    \end{minipage}\\*
    \begin{minipage}[t][4cm][t]{0.46\textwidth}
      \footnotesize
      \textbf{Q:} Is there another green rubber cube that has the same size as the green matte cube? \\*
      \textbf{A:} no \\*
      \textbf{Q-type:} exist \\*
      \textbf{Size:} 10 \\*[6pt]
    \end{minipage}\hspace{1mm}
    \begin{minipage}[t][4cm][t]{0.46\textwidth}
      \footnotesize
      \textbf{Q:} Is the large matte thing the same shape as the big red object? \\*
      \textbf{A:} yes \\*
      \textbf{Q-type:} equal\_shape \\*
      \textbf{Size:} 11 \\*[6pt]
    \end{minipage}
  \end{minipage}
  \hspace{1mm}
  \begin{minipage}{0.32\textwidth}
    \includegraphics[width=\textwidth]{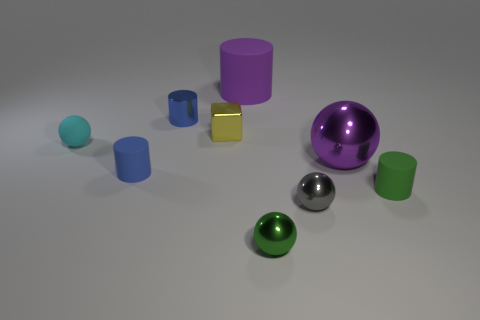}
    \begin{minipage}[t][4cm][t]{0.46\textwidth}
      \footnotesize
      \textbf{Q:} There is a tiny rubber thing that is the same color as the metal cylinder; what shape is it? \\*
      \textbf{A:} cylinder \\*
      \textbf{Q-type:} query\_shape \\*
      \textbf{Size:} 9 \\*[6pt]
    \end{minipage}\hspace{1mm}
    \begin{minipage}[t][4cm][t]{0.46\textwidth}
      \footnotesize
      \textbf{Q:} What is the shape of the tiny green thing that is made of the same material as the large cylinder? \\*
      \textbf{A:} cylinder \\*
      \textbf{Q-type:} query\_shape \\*
      \textbf{Size:} 9 \\*[6pt]
    \end{minipage}\\*
    \begin{minipage}[t][4cm][t]{0.46\textwidth}
      \footnotesize
      \textbf{Q:} Do the blue metallic object and the green metal thing have the same shape? \\*
      \textbf{A:} no \\*
      \textbf{Q-type:} equal\_shape \\*
      \textbf{Size:} 11 \\*[6pt]
    \end{minipage}\hspace{1mm}
    \begin{minipage}[t][4cm][t]{0.46\textwidth}
      \footnotesize
      \textbf{Q:} The big matte thing is what color? \\*
      \textbf{A:} purple \\*
      \textbf{Q-type:} query\_color \\*
      \textbf{Size:} 5 \\*[6pt]
    \end{minipage}
  \end{minipage}
  \hspace{1mm}
  \begin{minipage}{0.32\textwidth}
    \includegraphics[width=\textwidth]{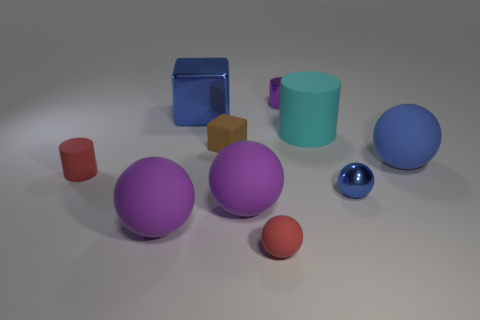}
    \begin{minipage}[t][4cm][t]{0.46\textwidth}
      \footnotesize
      \textbf{Q:} There is a small ball that is made of the same material as the large block; what color is it? \\*
      \textbf{A:} blue \\*
      \textbf{Q-type:} query\_color \\*
      \textbf{Size:} 9 \\*[6pt]
    \end{minipage}\hspace{1mm}
    \begin{minipage}[t][4cm][t]{0.46\textwidth}
      \footnotesize
      \textbf{Q:} Is the size of the red rubber sphere the same as the purple metal thing? \\*
      \textbf{A:} yes \\*
      \textbf{Q-type:} equal\_size \\*
      \textbf{Size:} 12 \\*[6pt]
    \end{minipage}\\*
    \begin{minipage}[t][4cm][t]{0.46\textwidth}
      \footnotesize
      \textbf{Q:} What is the material of the purple cylinder? \\*
      \textbf{A:} metal \\*
      \textbf{Q-type:} query\_material \\*
      \textbf{Size:} 5 \\*[6pt]
    \end{minipage}\hspace{1mm}
    \begin{minipage}[t][4cm][t]{0.46\textwidth}
      \footnotesize
      \textbf{Q:} There is a blue ball that is the same size as the brown thing; what material is it? \\*
      \textbf{A:} metal \\*
      \textbf{Q-type:} query\_material \\*
      \textbf{Size:} 8 \\*[6pt]
    \end{minipage}
  \end{minipage}
  \begin{minipage}{0.32\textwidth}
    \includegraphics[width=\textwidth]{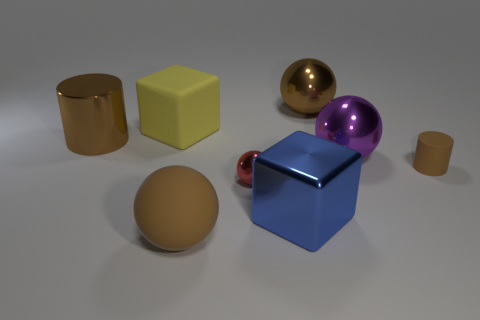}
    \begin{minipage}[t][4cm][t]{0.46\textwidth}
      \footnotesize
      \textbf{Q:} How many small spheres are the same color as the big rubber cube? \\*
      \textbf{A:} 0 \\*
      \textbf{Q-type:} count \\*
      \textbf{Size:} 9 \\*[6pt]
    \end{minipage}\hspace{1mm}
    \begin{minipage}[t][4cm][t]{0.46\textwidth}
      \footnotesize
      \textbf{Q:} Is the tiny ball made of the same material as the large purple ball? \\*
      \textbf{A:} yes \\*
      \textbf{Q-type:} equal\_material \\*
      \textbf{Size:} 12 \\*[6pt]
    \end{minipage}\\*
    \begin{minipage}[t][4cm][t]{0.46\textwidth}
      \footnotesize
      \textbf{Q:} There is a large cube that is right of the red sphere; what number of large yellow things are on the right side of it? \\*
      \textbf{A:} 0 \\*
      \textbf{Q-type:} count \\*
      \textbf{Size:} 12 \\*[6pt]
    \end{minipage}\hspace{1mm}
    \begin{minipage}[t][4cm][t]{0.46\textwidth}
      \footnotesize
      \textbf{Q:} Do the matte cylinder and the purple shiny object have the same size? \\*
      \textbf{A:} no \\*
      \textbf{Q-type:} equal\_size \\*
      \textbf{Size:} 11 \\*[6pt]
    \end{minipage}
  \end{minipage}
  \hspace{1mm}
  \begin{minipage}{0.32\textwidth}
    \includegraphics[width=\textwidth]{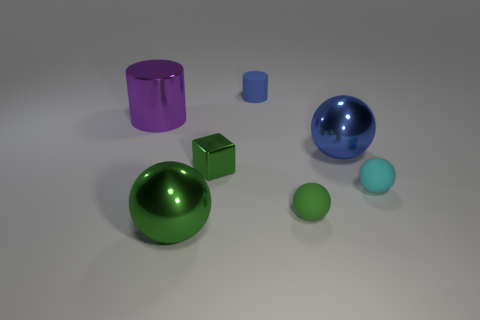}
    \begin{minipage}[t][4cm][t]{0.46\textwidth}
      \footnotesize
      \textbf{Q:} How many small green things are behind the green rubber sphere in front of the blue thing that is in front of the large purple metal cylinder? \\*
      \textbf{A:} 1 \\*
      \textbf{Q-type:} count \\*
      \textbf{Size:} 18 \\*[6pt]
    \end{minipage}\hspace{1mm}
    \begin{minipage}[t][4cm][t]{0.46\textwidth}
      \footnotesize
      \textbf{Q:} The cylinder that is the same size as the blue metallic sphere is what color? \\*
      \textbf{A:} purple \\*
      \textbf{Q-type:} query\_color \\*
      \textbf{Size:} 9 \\*[6pt]
    \end{minipage}\\*
    \begin{minipage}[t][4cm][t]{0.46\textwidth}
      \footnotesize
      \textbf{Q:} What is the size of the green metal object right of the large ball that is on the left side of the big blue metal sphere? \\*
      \textbf{A:} small \\*
      \textbf{Q-type:} query\_size \\*
      \textbf{Size:} 15 \\*[6pt]
    \end{minipage}\hspace{1mm}
    \begin{minipage}[t][4cm][t]{0.46\textwidth}
      \footnotesize
      \textbf{Q:} There is a metallic ball that is the same color as the tiny cylinder; what size is it? \\*
      \textbf{A:} large \\*
      \textbf{Q-type:} query\_size \\*
      \textbf{Size:} 9 \\*[6pt]
    \end{minipage}
  \end{minipage}
  \hspace{1mm}
  \begin{minipage}{0.32\textwidth}
    \includegraphics[width=\textwidth]{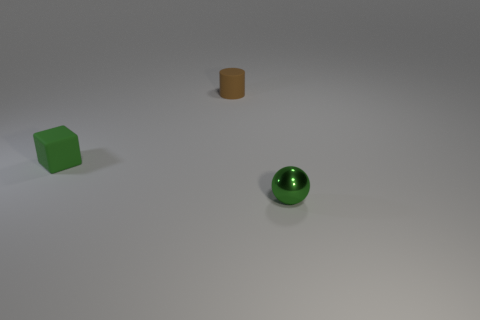}
    \begin{minipage}[t][4cm][t]{0.46\textwidth}
      \footnotesize
      \textbf{Q:} What is the color of the matte thing that is left of the thing behind the tiny green thing behind the tiny shiny sphere? \\*
      \textbf{A:} green \\*
      \textbf{Q-type:} query\_color \\*
      \textbf{Size:} 15 \\*[6pt]
    \end{minipage}\hspace{1mm}
    \begin{minipage}[t][4cm][t]{0.46\textwidth}
      \footnotesize
      \textbf{Q:} There is a thing in front of the tiny block; is its color the same as the matte object in front of the tiny brown rubber thing? \\*
      \textbf{A:} yes \\*
      \textbf{Q-type:} equal\_color \\*
      \textbf{Size:} 17 \\*[6pt]
    \end{minipage}\\*
    \begin{minipage}[t][4cm][t]{0.46\textwidth}
      \footnotesize
      \textbf{Q:} The green thing behind the small green thing on the right side of the brown matte object is what shape? \\*
      \textbf{A:} cube \\*
      \textbf{Q-type:} query\_shape \\*
      \textbf{Size:} 12 \\*[6pt]
    \end{minipage}\hspace{1mm}
    \begin{minipage}[t][4cm][t]{0.46\textwidth}
      \footnotesize
      \textbf{Q:} Is there a green matte cube that has the same size as the shiny thing? \\*
      \textbf{A:} yes \\*
      \textbf{Q-type:} exist \\*
      \textbf{Size:} 8 \\*[6pt]
    \end{minipage}
  \end{minipage}
\end{figure*}
\begin{figure*}
  \begin{minipage}{0.32\textwidth}
    \includegraphics[width=\textwidth]{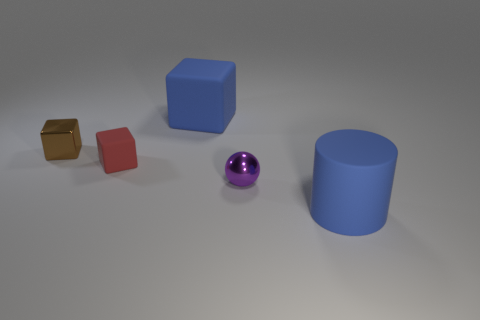}
    \begin{minipage}[t][4cm][t]{0.46\textwidth}
      \footnotesize
      \textbf{Q:} Are there more brown shiny objects behind the large rubber cylinder than gray blocks? \\*
      \textbf{A:} yes \\*
      \textbf{Q-type:} greater\_than \\*
      \textbf{Size:} 14 \\*[6pt]
    \end{minipage}\hspace{1mm}
    \begin{minipage}[t][4cm][t]{0.46\textwidth}
      \footnotesize
      \textbf{Q:} What color is the matte object to the right of the large block? \\*
      \textbf{A:} blue \\*
      \textbf{Q-type:} query\_color \\*
      \textbf{Size:} 8 \\*[6pt]
    \end{minipage}\\*
    \begin{minipage}[t][4cm][t]{0.46\textwidth}
      \footnotesize
      \textbf{Q:} Do the blue cube and the cylinder have the same size? \\*
      \textbf{A:} yes \\*
      \textbf{Q-type:} equal\_size \\*
      \textbf{Size:} 10 \\*[6pt]
    \end{minipage}\hspace{1mm}
    \begin{minipage}[t][4cm][t]{0.46\textwidth}
      \footnotesize
      \textbf{Q:} The ball has what size? \\*
      \textbf{A:} small \\*
      \textbf{Q-type:} query\_size \\*
      \textbf{Size:} 4 \\*[6pt]
    \end{minipage}
  \end{minipage}
  \hspace{1mm}
  \begin{minipage}{0.32\textwidth}
    \includegraphics[width=\textwidth]{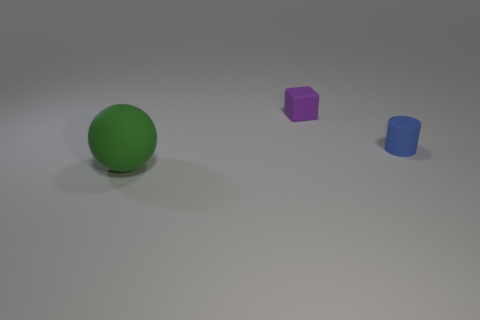}
    \begin{minipage}[t][4cm][t]{0.46\textwidth}
      \footnotesize
      \textbf{Q:} Are there any other things that are the same shape as the large green thing? \\*
      \textbf{A:} no \\*
      \textbf{Q-type:} exist \\*
      \textbf{Size:} 6 \\*[6pt]
    \end{minipage}\hspace{1mm}
    \begin{minipage}[t][4cm][t]{0.46\textwidth}
      \footnotesize
      \textbf{Q:} The matte thing that is both in front of the purple cube and to the left of the blue rubber cylinder is what color? \\*
      \textbf{A:} green \\*
      \textbf{Q-type:} query\_color \\*
      \textbf{Size:} 15 \\*[6pt]
    \end{minipage}\\*
    \begin{minipage}[t][4cm][t]{0.46\textwidth}
      \footnotesize
      \textbf{Q:} Are there fewer small rubber cylinders in front of the green ball than purple cylinders? \\*
      \textbf{A:} no \\*
      \textbf{Q-type:} less\_than \\*
      \textbf{Size:} 14 \\*[6pt]
    \end{minipage}\hspace{1mm}
    \begin{minipage}[t][4cm][t]{0.46\textwidth}
      \footnotesize
      \textbf{Q:} What number of other objects are there of the same shape as the large rubber object? \\*
      \textbf{A:} 0 \\*
      \textbf{Q-type:} count \\*
      \textbf{Size:} 6 \\*[6pt]
    \end{minipage}
  \end{minipage}
  \hspace{1mm}
  \begin{minipage}{0.32\textwidth}
    \includegraphics[width=\textwidth]{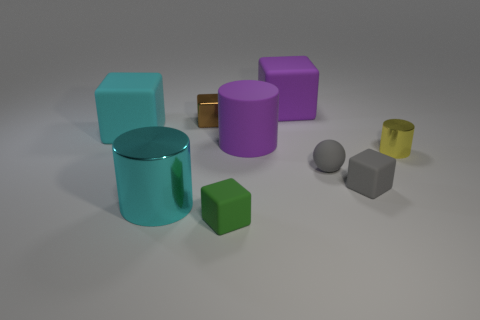}
    \begin{minipage}[t][4cm][t]{0.46\textwidth}
      \footnotesize
      \textbf{Q:} What number of cubes are the same color as the rubber ball? \\*
      \textbf{A:} 1 \\*
      \textbf{Q-type:} count \\*
      \textbf{Size:} 7 \\*[6pt]
    \end{minipage}\hspace{1mm}
    \begin{minipage}[t][4cm][t]{0.46\textwidth}
      \footnotesize
      \textbf{Q:} Is the big cyan metal thing the same shape as the brown thing? \\*
      \textbf{A:} no \\*
      \textbf{Q-type:} equal\_shape \\*
      \textbf{Size:} 11 \\*[6pt]
    \end{minipage}\\*
    \begin{minipage}[t][4cm][t]{0.46\textwidth}
      \footnotesize
      \textbf{Q:} What shape is the tiny metal thing behind the large block in front of the tiny brown block? \\*
      \textbf{A:} cube \\*
      \textbf{Q-type:} query\_shape \\*
      \textbf{Size:} 14 \\*[6pt]
    \end{minipage}\hspace{1mm}
    \begin{minipage}[t][4cm][t]{0.46\textwidth}
      \footnotesize
      \textbf{Q:} Is the size of the cyan cube the same as the metal cylinder that is behind the cyan cylinder? \\*
      \textbf{A:} no \\*
      \textbf{Q-type:} equal\_size \\*
      \textbf{Size:} 15 \\*[6pt]
    \end{minipage}
  \end{minipage}
  \begin{minipage}{0.32\textwidth}
    \includegraphics[width=\textwidth]{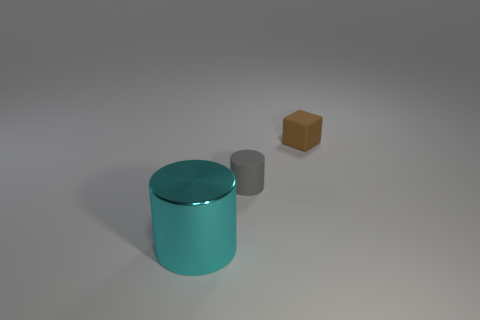}
    \begin{minipage}[t][4cm][t]{0.46\textwidth}
      \footnotesize
      \textbf{Q:} There is a tiny brown rubber thing; is its shape the same as the thing that is in front of the small matte cylinder? \\*
      \textbf{A:} no \\*
      \textbf{Q-type:} equal\_shape \\*
      \textbf{Size:} 15 \\*[6pt]
    \end{minipage}\hspace{1mm}
    \begin{minipage}[t][4cm][t]{0.46\textwidth}
      \footnotesize
      \textbf{Q:} What number of yellow rubber things are the same size as the gray thing? \\*
      \textbf{A:} 0 \\*
      \textbf{Q-type:} count \\*
      \textbf{Size:} 7 \\*[6pt]
    \end{minipage}\\*
    \begin{minipage}[t][4cm][t]{0.46\textwidth}
      \footnotesize
      \textbf{Q:} What number of tiny brown rubber objects are behind the rubber object that is on the right side of the cylinder on the right side of the big cyan cylinder? \\*
      \textbf{A:} 0 \\*
      \textbf{Q-type:} count \\*
      \textbf{Size:} 16 \\*[6pt]
    \end{minipage}\hspace{1mm}
    \begin{minipage}[t][4cm][t]{0.46\textwidth}
      \footnotesize
      \textbf{Q:} Are there an equal number of brown matte objects that are right of the tiny brown object and gray rubber cylinders that are to the left of the large cyan shiny cylinder? \\*
      \textbf{A:} yes \\*
      \textbf{Q-type:} equal\_integer \\*
      \textbf{Size:} 20 \\*[6pt]
    \end{minipage}
  \end{minipage}
  \hspace{1mm}
  \begin{minipage}{0.32\textwidth}
    \includegraphics[width=\textwidth]{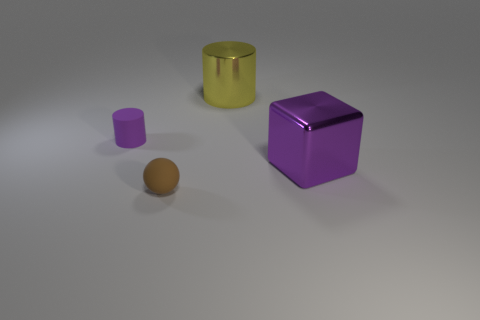}
    \begin{minipage}[t][4cm][t]{0.46\textwidth}
      \footnotesize
      \textbf{Q:} Are there any rubber things that have the same size as the yellow metallic cylinder? \\*
      \textbf{A:} no \\*
      \textbf{Q-type:} exist \\*
      \textbf{Size:} 8 \\*[6pt]
    \end{minipage}\hspace{1mm}
    \begin{minipage}[t][4cm][t]{0.46\textwidth}
      \footnotesize
      \textbf{Q:} What is the size of the brown sphere? \\*
      \textbf{A:} small \\*
      \textbf{Q-type:} query\_size \\*
      \textbf{Size:} 5 \\*[6pt]
    \end{minipage}\\*
    \begin{minipage}[t][4cm][t]{0.46\textwidth}
      \footnotesize
      \textbf{Q:} What number of yellow metal objects are the same size as the metallic cube? \\*
      \textbf{A:} 1 \\*
      \textbf{Q-type:} count \\*
      \textbf{Size:} 8 \\*[6pt]
    \end{minipage}\hspace{1mm}
    \begin{minipage}[t][4cm][t]{0.46\textwidth}
      \footnotesize
      \textbf{Q:} Is the number of purple matte cylinders behind the large purple thing less than the number of tiny rubber objects in front of the big cylinder? \\*
      \textbf{A:} yes \\*
      \textbf{Q-type:} less\_than \\*
      \textbf{Size:} 18 \\*[6pt]
    \end{minipage}
  \end{minipage}
  \hspace{1mm}
  \begin{minipage}{0.32\textwidth}
    \includegraphics[width=\textwidth]{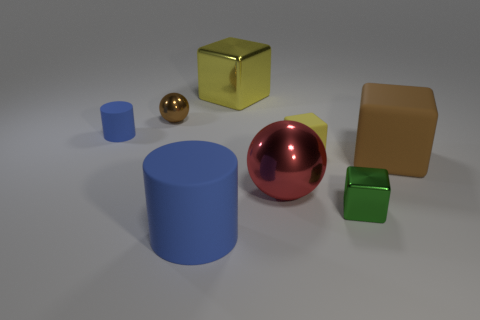}
    \begin{minipage}[t][4cm][t]{0.46\textwidth}
      \footnotesize
      \textbf{Q:} What is the shape of the shiny thing that is behind the small blue rubber object and to the right of the tiny brown thing? \\*
      \textbf{A:} cube \\*
      \textbf{Q-type:} query\_shape \\*
      \textbf{Size:} 15 \\*[6pt]
    \end{minipage}\hspace{1mm}
    \begin{minipage}[t][4cm][t]{0.46\textwidth}
      \footnotesize
      \textbf{Q:} What is the shape of the blue rubber object in front of the brown object to the right of the big metallic block? \\*
      \textbf{A:} cylinder \\*
      \textbf{Q-type:} query\_shape \\*
      \textbf{Size:} 13 \\*[6pt]
    \end{minipage}\\*
    \begin{minipage}[t][4cm][t]{0.46\textwidth}
      \footnotesize
      \textbf{Q:} Does the large red object have the same shape as the large blue thing? \\*
      \textbf{A:} no \\*
      \textbf{Q-type:} equal\_shape \\*
      \textbf{Size:} 11 \\*[6pt]
    \end{minipage}\hspace{1mm}
    \begin{minipage}[t][4cm][t]{0.46\textwidth}
      \footnotesize
      \textbf{Q:} There is a large cylinder that is the same color as the tiny cylinder; what is it made of? \\*
      \textbf{A:} rubber \\*
      \textbf{Q-type:} query\_material \\*
      \textbf{Size:} 9 \\*[6pt]
    \end{minipage}
  \end{minipage}
\end{figure*}
\begin{figure*}
  \begin{minipage}{0.32\textwidth}
    \includegraphics[width=\textwidth]{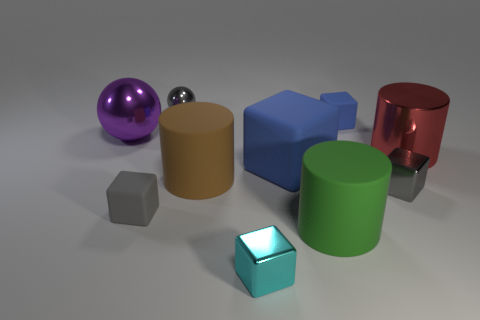}
    \begin{minipage}[t][4cm][t]{0.46\textwidth}
      \footnotesize
      \textbf{Q:} There is a red shiny thing right of the purple metal sphere; what is its shape? \\*
      \textbf{A:} cylinder \\*
      \textbf{Q-type:} query\_shape \\*
      \textbf{Size:} 10 \\*[6pt]
    \end{minipage}\hspace{1mm}
    \begin{minipage}[t][4cm][t]{0.46\textwidth}
      \footnotesize
      \textbf{Q:} What number of purple shiny things are there? \\*
      \textbf{A:} 1 \\*
      \textbf{Q-type:} count \\*
      \textbf{Size:} 4 \\*[6pt]
    \end{minipage}\\*
    \begin{minipage}[t][4cm][t]{0.46\textwidth}
      \footnotesize
      \textbf{Q:} Are the red object and the cyan cube made of the same material? \\*
      \textbf{A:} yes \\*
      \textbf{Q-type:} equal\_material \\*
      \textbf{Size:} 10 \\*[6pt]
    \end{minipage}\hspace{1mm}
    \begin{minipage}[t][4cm][t]{0.46\textwidth}
      \footnotesize
      \textbf{Q:} Are there more metallic objects that are right of the large red shiny cylinder than gray matte objects? \\*
      \textbf{A:} no \\*
      \textbf{Q-type:} greater\_than \\*
      \textbf{Size:} 14 \\*[6pt]
    \end{minipage}
  \end{minipage}
  \hspace{1mm}
  \begin{minipage}{0.32\textwidth}
    \includegraphics[width=\textwidth]{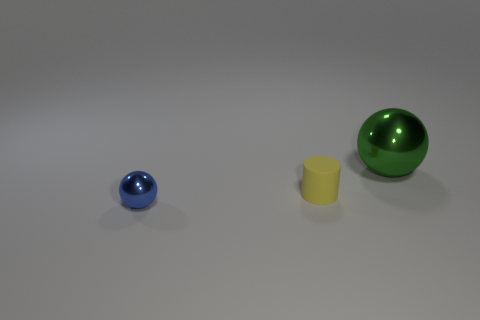}
    \begin{minipage}[t][4cm][t]{0.46\textwidth}
      \footnotesize
      \textbf{Q:} Does the small ball have the same color as the small cylinder in front of the big sphere? \\*
      \textbf{A:} no \\*
      \textbf{Q-type:} equal\_color \\*
      \textbf{Size:} 15 \\*[6pt]
    \end{minipage}\hspace{1mm}
    \begin{minipage}[t][4cm][t]{0.46\textwidth}
      \footnotesize
      \textbf{Q:} How many other things are the same size as the green sphere? \\*
      \textbf{A:} 0 \\*
      \textbf{Q-type:} count \\*
      \textbf{Size:} 6 \\*[6pt]
    \end{minipage}\\*
    \begin{minipage}[t][4cm][t]{0.46\textwidth}
      \footnotesize
      \textbf{Q:} How many blocks are yellow rubber things or large green shiny things? \\*
      \textbf{A:} 0 \\*
      \textbf{Q-type:} count \\*
      \textbf{Size:} 10 \\*[6pt]
    \end{minipage}\hspace{1mm}
    \begin{minipage}[t][4cm][t]{0.46\textwidth}
      \footnotesize
      \textbf{Q:} There is a shiny object in front of the yellow cylinder; is it the same shape as the yellow thing? \\*
      \textbf{A:} no \\*
      \textbf{Q-type:} equal\_shape \\*
      \textbf{Size:} 13 \\*[6pt]
    \end{minipage}
  \end{minipage}
  \hspace{1mm}
  \begin{minipage}{0.32\textwidth}
    \includegraphics[width=\textwidth]{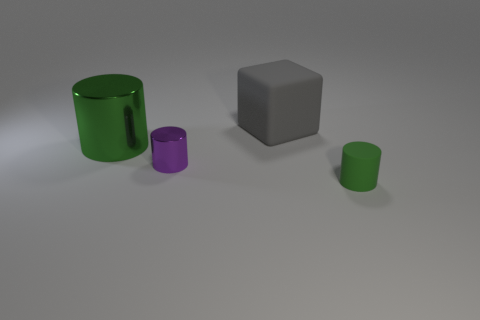}
    \begin{minipage}[t][4cm][t]{0.46\textwidth}
      \footnotesize
      \textbf{Q:} There is a object that is behind the big green metal cylinder; what is its material? \\*
      \textbf{A:} rubber \\*
      \textbf{Q-type:} query\_material \\*
      \textbf{Size:} 9 \\*[6pt]
    \end{minipage}\hspace{1mm}
    \begin{minipage}[t][4cm][t]{0.46\textwidth}
      \footnotesize
      \textbf{Q:} How big is the gray thing? \\*
      \textbf{A:} large \\*
      \textbf{Q-type:} query\_size \\*
      \textbf{Size:} 4 \\*[6pt]
    \end{minipage}\\*
    \begin{minipage}[t][4cm][t]{0.46\textwidth}
      \footnotesize
      \textbf{Q:} Do the green object behind the tiny green matte cylinder and the small purple object have the same material? \\*
      \textbf{A:} yes \\*
      \textbf{Q-type:} equal\_material \\*
      \textbf{Size:} 16 \\*[6pt]
    \end{minipage}\hspace{1mm}
    \begin{minipage}[t][4cm][t]{0.46\textwidth}
      \footnotesize
      \textbf{Q:} What number of rubber blocks are there? \\*
      \textbf{A:} 1 \\*
      \textbf{Q-type:} count \\*
      \textbf{Size:} 4 \\*[6pt]
    \end{minipage}
  \end{minipage}
  \begin{minipage}{0.32\textwidth}
    \includegraphics[width=\textwidth]{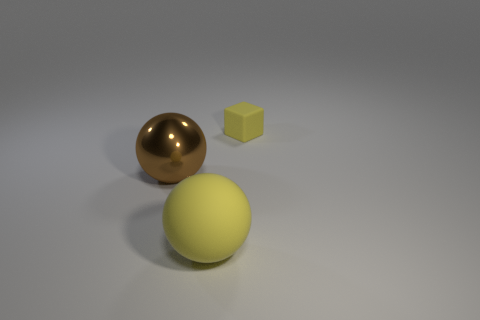}
    \begin{minipage}[t][4cm][t]{0.46\textwidth}
      \footnotesize
      \textbf{Q:} How big is the yellow thing behind the brown shiny thing? \\*
      \textbf{A:} small \\*
      \textbf{Q-type:} query\_size \\*
      \textbf{Size:} 8 \\*[6pt]
    \end{minipage}\hspace{1mm}
    \begin{minipage}[t][4cm][t]{0.46\textwidth}
      \footnotesize
      \textbf{Q:} Is the size of the brown metal ball the same as the yellow thing that is behind the yellow rubber sphere? \\*
      \textbf{A:} no \\*
      \textbf{Q-type:} equal\_size \\*
      \textbf{Size:} 16 \\*[6pt]
    \end{minipage}\\*
    \begin{minipage}[t][4cm][t]{0.46\textwidth}
      \footnotesize
      \textbf{Q:} Are there fewer small yellow things to the left of the large yellow matte ball than large brown objects? \\*
      \textbf{A:} yes \\*
      \textbf{Q-type:} less\_than \\*
      \textbf{Size:} 15 \\*[6pt]
    \end{minipage}\hspace{1mm}
    \begin{minipage}[t][4cm][t]{0.46\textwidth}
      \footnotesize
      \textbf{Q:} What material is the other thing that is the same shape as the brown thing? \\*
      \textbf{A:} rubber \\*
      \textbf{Q-type:} query\_material \\*
      \textbf{Size:} 6 \\*[6pt]
    \end{minipage}
  \end{minipage}
  \hspace{1mm}
  \begin{minipage}{0.32\textwidth}
    \includegraphics[width=\textwidth]{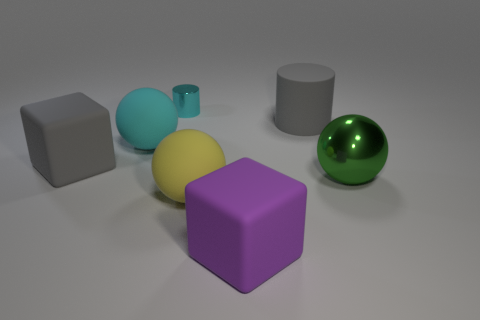}
    \begin{minipage}[t][4cm][t]{0.46\textwidth}
      \footnotesize
      \textbf{Q:} How many things are either green things or matte cubes behind the green ball? \\*
      \textbf{A:} 2 \\*
      \textbf{Q-type:} count \\*
      \textbf{Size:} 11 \\*[6pt]
    \end{minipage}\hspace{1mm}
    \begin{minipage}[t][4cm][t]{0.46\textwidth}
      \footnotesize
      \textbf{Q:} There is a cyan object that is to the right of the cyan rubber sphere; is its size the same as the gray rubber cylinder? \\*
      \textbf{A:} no \\*
      \textbf{Q-type:} equal\_size \\*
      \textbf{Size:} 16 \\*[6pt]
    \end{minipage}\\*
    \begin{minipage}[t][4cm][t]{0.46\textwidth}
      \footnotesize
      \textbf{Q:} There is a sphere to the right of the large yellow ball; what material is it? \\*
      \textbf{A:} metal \\*
      \textbf{Q-type:} query\_material \\*
      \textbf{Size:} 9 \\*[6pt]
    \end{minipage}\hspace{1mm}
    \begin{minipage}[t][4cm][t]{0.46\textwidth}
      \footnotesize
      \textbf{Q:} Are there the same number of tiny cylinders that are behind the cyan metal object and purple blocks right of the gray cube? \\*
      \textbf{A:} no \\*
      \textbf{Q-type:} equal\_integer \\*
      \textbf{Size:} 17 \\*[6pt]
    \end{minipage}
  \end{minipage}
  \hspace{1mm}
  \begin{minipage}{0.32\textwidth}
    \includegraphics[width=\textwidth]{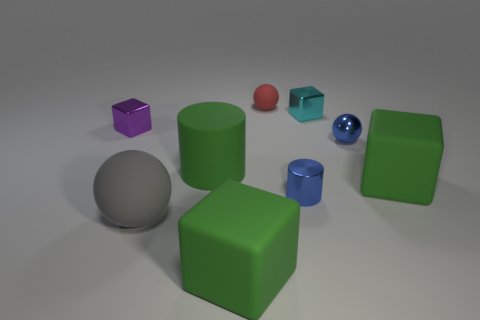}
    \begin{minipage}[t][4cm][t]{0.46\textwidth}
      \footnotesize
      \textbf{Q:} What color is the rubber ball in front of the metal cube to the left of the matte cube left of the blue metallic sphere? \\*
      \textbf{A:} gray \\*
      \textbf{Q-type:} query\_color \\*
      \textbf{Size:} 18 \\*[6pt]
    \end{minipage}\hspace{1mm}
    \begin{minipage}[t][4cm][t]{0.46\textwidth}
      \footnotesize
      \textbf{Q:} What shape is the cyan shiny thing that is the same size as the red matte object? \\*
      \textbf{A:} cube \\*
      \textbf{Q-type:} query\_shape \\*
      \textbf{Size:} 9 \\*[6pt]
    \end{minipage}\\*
    \begin{minipage}[t][4cm][t]{0.46\textwidth}
      \footnotesize
      \textbf{Q:} There is a green cylinder on the right side of the big gray ball; does it have the same size as the ball that is behind the metal ball? \\*
      \textbf{A:} no \\*
      \textbf{Q-type:} equal\_size \\*
      \textbf{Size:} 19 \\*[6pt]
    \end{minipage}\hspace{1mm}
    \begin{minipage}[t][4cm][t]{0.46\textwidth}
      \footnotesize
      \textbf{Q:} What is the size of the green block behind the big gray matte sphere? \\*
      \textbf{A:} large \\*
      \textbf{Q-type:} query\_size \\*
      \textbf{Size:} 11 \\*[6pt]
    \end{minipage}
  \end{minipage}
\end{figure*}

\clearpage
\pagebreak
\textbf{Acknowledgments}
We thank Deepak Pathak, Piotr Doll{\'a}r, Ranjay Krishna, Animesh Garg, and
Danfei Xu for helpful comments and discussion.

{\small
\bibliographystyle{ieee}
\bibliography{biblio}
}

\end{document}